\documentclass{ieeeaccess}

\usepackage{cite}
\usepackage{amsmath,amssymb,amsfonts}
\usepackage{algorithmic}
\usepackage{graphicx}
\usepackage{textcomp}
\usepackage{multirow}
\usepackage{tabularx}
\usepackage[caption=false]{subfig}

\def\BibTeX{{\rm B\kern-.05em{\sc i\kern-.025em b}\kern-.08em
    T\kern-.1667em\lower.7ex\hbox{E}\kern-.125emX}}

\newcolumntype{Y}{>{\centering\arraybackslash}X}

\begin{document}
\history{Date of publication xxxx 00, 0000, date of current version xxxx 00, 0000.}
\doi{10.1109/ACCESS.2024.3492703}

\title{Shape2.5D: A Dataset of Texture-less Surfaces for Depth and Normals Estimation}
\author{\uppercase{Muhammad Saif Ullah Khan}\authorrefmark{1,2}, \uppercase{Sankalp Sinha}\authorrefmark{1,2},
\uppercase{Didier Stricker\authorrefmark{1,2}, Marcus Liwicki\authorrefmark{3}, and Muhammad Zeshan Afzal}.\authorrefmark{1,2}}
\address[1]{Rhineland-Palatinate Technical University of Kaiserslautern-Landau (RPTU), 67663 Kaiserslautern, Germany}
\address[2]{German Research Center for Artificial Intelligence (DFKI), 67663 Kaiserslautern, Germany}
\address[3]{Luleå University of Technology, 97187 Luleå, Sweden}
% \tfootnote{This work was supported in part by the Horizon Europe project LUMINOUS under Grant Agreement 101135724.}

\markboth
{Khan \headeretal: A Dataset of Texture-less Surfaces for Depth and Normals Estimation}
{Khan \headeretal: A Dataset of Texture-less Surfaces for Depth and Normals Estimation}

\corresp{Corresponding author: Muhammad Saif Ullah Khan (e-mail: muhammad\_saif\_ullah.khan@dfki.de).}

\begin{abstract}
Reconstructing texture-less surfaces poses unique challenges in computer vision, primarily due to the lack of specialized datasets that cater to the nuanced needs of depth and normals estimation in the absence of textural information. We introduce ”Shape2.5D,” a novel, large-scale dataset designed to address this gap. Comprising 1.17 million frames spanning over 39,772 3D models and 48 unique objects, our dataset provides depth and surface normal maps for texture-less object reconstruction. The proposed dataset includes synthetic images rendered with 3D modeling software to simulate various lighting conditions and viewing angles. It also includes a real-world subset comprising 4,672 frames captured with a depth camera. Our comprehensive benchmarks demonstrate the dataset’s ability to support the development of algorithms that robustly estimate depth and normals from RGB images and perform voxel reconstruction. Our open-source data generation pipeline allows the dataset to be extended and adapted for future research. The dataset is publicly available at \underline{https://github.com/saifkhichi96/Shape25D}.
\end{abstract}


\titlepgskip=-15pt

\maketitle

\Figure[!t]()[width=\linewidth]{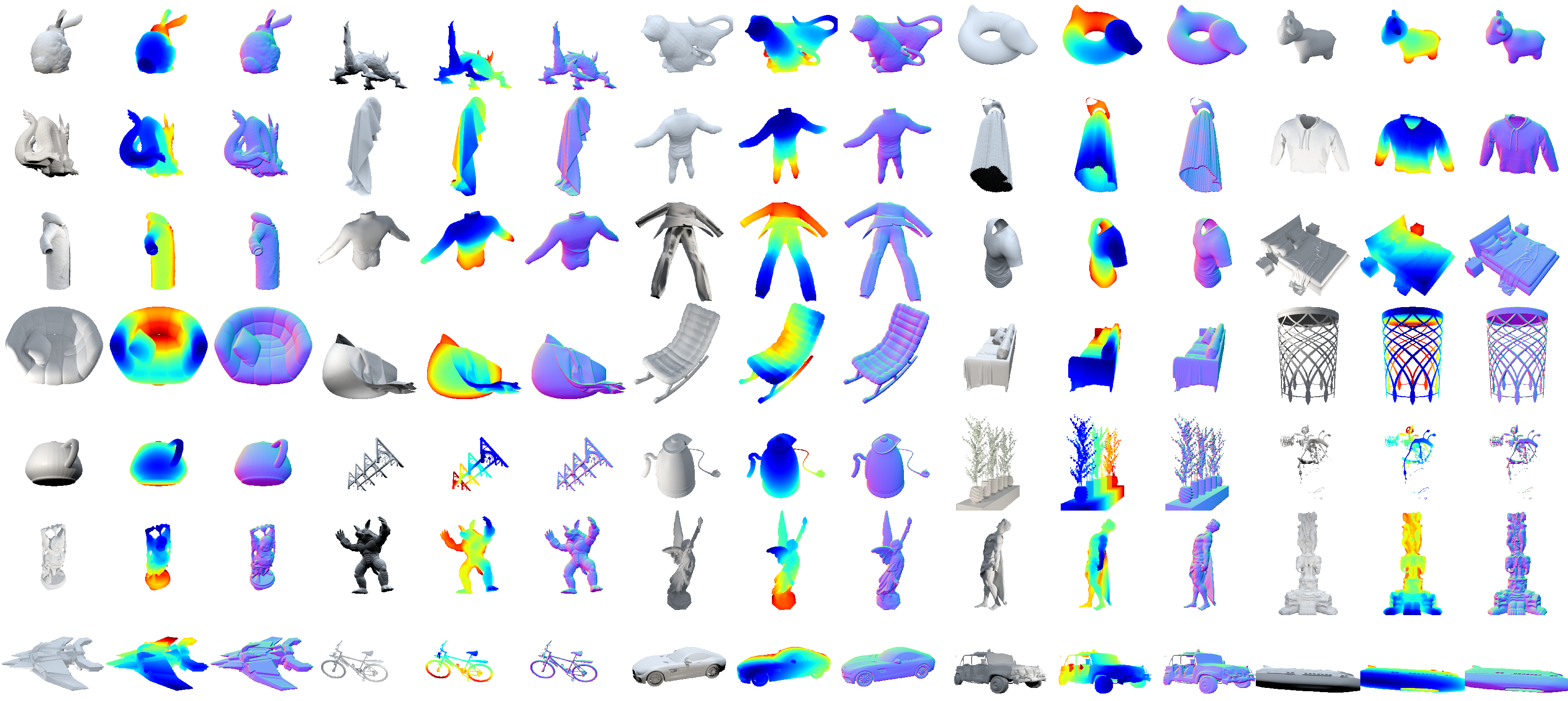}
{\textbf{The Shape2.5D dataset} comprises two synthetic and one real-world component, providing texture-less images, depth maps, and normal maps. We show samples from the Synthetic (A) set with 35 common items from categories including animals, clothing, furniture, statues, vehicles, and miscellaneous objects.\label{fig:data_synthetic}}

\section{Introduction}
\label{sec:intro}
\PARstart{S}{ignificant} advances have been made in 3D surface reconstruction in recent years~\cite{wang2018pixel2mesh, bednarik2018learning, golyanik2018hdm, shimada2019ismo, tsoli2019patch, li2021vrvt}. Traditional works~\cite{ley2016reconstructing,wang2016template,hafeez2017image,ahmadabadian2019automatic,santovsi2019evaluation,fan20213d,cheng2021multi,yang2021practical} have also explored the reconstruction of weakly-textured surfaces. Textures refer to the distinctive patterns or visual details present on surfaces, which aid in identifying features and reconstructing shapes. In contrast, texture-less objects lack these visual details, making reconstruction significantly more challenging. However, these surfaces have received limited attention in deep learning~\cite{bednarik2018learning, tsoli2019patch,golyanik2018hdm,ye2023self}. From industrial components~\cite{hodan2017t} and planar surfaces in indoor scenes~\cite{ley2016reconstructing,hafeez2017image,ye2023self}, to biological tissues in medical imaging~\cite{widya2019whole,boven2020diagnostic,fan20213d}, texture-less surfaces are everywhere. Therefore, the need for effective reconstruction methods is critical and practical.

Deep learning methods rely heavily on data, making it a core foundation of their effectiveness~\cite{sun2017revisiting}. However, very few shape datasets target texture-less objects~\cite{hodan2017t,bednarik2018learning}. None of these are large-scale like their textured counterparts~\cite{chang2015shapenet,sun2018pix3d,xie2020pix2vox++}. Additionally, the available texture-less datasets often only contain partial shape information, such as depth maps or surface normal maps. Full 3D ground truth in the form of point clouds, meshes, or voxel grids is required to train robust models. The scarcity of extensive, publicly available datasets for low-texture surfaces significantly impedes progress in this field~\cite{khan2022three}. This makes curating a comprehensive shape dataset of texture-less surfaces paramount. In light of these challenges, we present three key contributions:

\begin{itemize}
    \item We introduce a pioneering synthetic dataset of depth and normal maps of texture-less surfaces containing 48 unique objects (Figure~\ref{fig:data_synthetic}).
    \item A supplementary dataset of six texture-less objects in the real world with 4,345 frames from a depth camera.
    \item A comprehensive benchmark demonstrating the effectiveness of our dataset in depth estimation, surface normals estimation, and voxel reconstruction.
\end{itemize}

Moreover, we open-source our data generation and collection pipelines, enabling researchers to extend the dataset. By addressing the data scarcity issue head-on, we aim to enable advancements in texture-less surface reconstruction.

\section{Related Work}
\label{sec:literature}

\noindent \textbf{Depth and Normals Estimation} \hspace{2pt} Depth estimation~\cite{godard2019digging,costanzino2023learning,spencer2023monocular,gasperini2023robust,ming2021deep} determines the distance of each point in the image from the camera. On the other hand, surface normals estimation~\cite{lenssen2020deep,klasing2009comparison,wang2015designing} identifies the direction a surface faces at each point, which is vital for understanding the shape and contours of objects in the image. Together, these measurements are referred to as a 2.5D representation in the literature~\cite{wu2017marrnet}. A 2.5D representation combines depth and surface orientation information, providing more spatial details than a flat 2D image but lacking the full volumetric detail of a complete 3D model. This intermediary step toward full 3D reconstruction offers essential spatial information about the scene's structure.

\noindent \textbf{3D Reconstruction} \hspace{2pt} 3D reconstruction involves converting 2D image data into 3D models through various techniques~\cite{yuniarti2019review,maxim2021survey,khan2022three,mittal2022autosdf,wen20223d,alwala2022pre,tang2021skeletonnet,zhang2020training,xing2022few,xing2022semi,chen2023single,choy20163d,wang2018pixel2mesh,tiong20223d,yang2022fvor}. There are several ways of reconstructing an object in 3D. Some methods use point clouds~\cite{golyanik2018hdm,shimada2019ismo} to identify key points in space and recreate the object's surface. Other methods use voxels to represent the 3D volume in a grid~\cite{choy20163d,li2021vrvt,xing2022few,xing2022semi,tiong20223d,xie2019pix2vox,xie2020pix2vox++,shi20213d}. Additionally, some approaches reconstruct meshes that use vertices and interconnected faces to create a highly detailed representation of the object~\cite{wang2018pixel2mesh,yuan2021vanet,tang2021skeletonnet}.

\noindent \textbf{Texture-less Surfaces} \hspace{2pt} Traditional vision-based works have attempted to reconstruct texture-less surfaces using different methods.~\cite{cheng2021multi,santovsi2019evaluation} project patterns onto the surface,~\cite{hafeez2017image} uses a multi-camera setup to establish correspondences between pixels in different views of the surface, and~\cite{fan20213d} combines stereovision with Shape from Shading.~\cite{bednarik2018learning,tsoli2019patch} pioneered the texture-less object reconstruction in deep learning.~\cite{bednarik2018learning} also published a dataset of textureless clothes, which was used by some later works~\cite{tsoli2019patch,golyanik2018hdm}.

\noindent \textbf{Datasets for Surface Reconstruction} \hspace{2pt} Datasets like ShapeNet~\cite{chang2015shapenet}, Pix3D~\cite{sun2018pix3d}, and Things3D~\cite{xie2020pix2vox++} have significantly contributed to the progress in textured object reconstruction. Their significant size and variety have established benchmarks in the domain. However, texture-less objects have not received a similar focus, with datasets such as T-LESS~\cite{hodan2017t} and the work by~\cite{bednarik2018learning} offering limited diversity in terms of object types, number of shapes, and domain coverage. The T-LESS dataset targets 6D pose estimation of rigid, texture-less objects but comprises only 30 industrial objects. These were captured using structured light and time-of-flight RGB-D sensors, along with a high-resolution RGB camera, resulting in about 49k images. Although T-LESS provides two types of 3D models per object with accurate 6D pose annotations, its focus on a limited number of industry-related objects restricts its applicability across broader scenarios. Similarly, the dataset from~\cite{bednarik2018learning}, which was collected using a Microsoft Kinect Xbox 360 camera, contains only 26k samples of five deformable objects (clothes and paper). It focuses on the 3D reconstruction of deformable surfaces with uniform albedo under varied lighting conditions, including RGB images, normal maps, and depth maps. However, the small number of objects and limited diversity prevent comprehensive learning for broader use cases.

\begin{table}[ht]
    \centering
    \begin{tabularx}{\linewidth}{|l|p{2.3cm}|X|}
    \hline
    Subset & Object Types & Source \\
    \hline
    Synthetic (A) & 35 everyday objects from six natural scene categories & Rendered against a black background in a fixed 3D scene \\
    \hline
    Synthetic (B) & 13 common ShapeNet objects & Rendered in 579 diverse environments using HDRI lighting \\
    \hline
    Real-World & 4 clothing items and 2 household objects & Captured using a Microsoft Kinect sensor \\
    \hline
    \end{tabularx}
    \caption{Summary of dataset subsets, including object types and their respective rendering or collection methodologies.}
    \label{tab:dataset_summary}
\end{table}

Our dataset addresses these gaps by providing a more extensive collection of texture-less surfaces, featuring a larger variety of objects and detailed depth and normal maps. By expanding the diversity of shapes and covering a wider range of object types and conditions, we aim to encourage advancements in shape reconstruction from texture-less data, addressing the limitations of prior datasets.

\section{The Shape2.5D Dataset}
\label{sec:dataset}

In this section, we explain our dataset and its creation process. The dataset has two synthetic components rendered in Blender~\cite{community2018blender} and a real-world component obtained from a Microsoft Kinect V2 camera. These are summarized in Table~\ref{tab:dataset_summary}. We explain their creation process and composition in more detail in the following subsections.

\subsection{Synthetic Data}
\label{sec:dataset_synthetic}

\subsubsection{Synthetic (A) Subset}

This subset comprises 35 common objects, classified into six principal categories: \textit{animals}, \textit{clothing}, \textit{furniture}, \textit{statues}, \textit{vehicles}, and \textit{miscellaneous}, as detailed in Table~\ref{tab:data_synthetic}. Items such as clothing and furniture are frequently encountered in natural environments and tend to exhibit minimal or no textural attributes. Architectural sculptures are prevalent elements in numerous real-world contexts globally and are frequently characterized by a lack of color or significant texture. Our dataset includes several statues, featuring intricate, life-size 3D models that exhibit complex shapes and variations in depth, as depicted in Figure~\ref{fig:data_synthetic}.

\begin{table}[ht]
\setlength{\tabcolsep}{3pt}
\caption{\textbf{Summary of Synthetic (A) subset.} This dataset comprises 35 distinct objects, each represented by a single 3D model. It aims to offer thorough view coverage of these objects under different lighting conditions. We render each model from 360 degrees under 24 different setups, resulting in 216k samples. RGB images, depth maps, and surface normals are provided.}
\label{tab:data_splits}
\begin{tabularx}{\linewidth}{|X|ccc|}
\hline
Category (Objects) & Seq. & Views & Samples \\
\hline
\underline{animals} ({asian dragon}, {cats}, {duck}, {pig}, {stanford bunny}, {stanford dragon}) & 24 & 360 & 51,840 \\
\underline{clothing} ({cape}, {dress}, {hoodie}, {jacket}, {shirt}, {suit}, {tracksuit}, {tshirt}) & 24 & 360 & 69,120 \\
\underline{furniture} ({armchair}, {bed}, {chair}, {desk}, {rocking chair}, {sofa}) & 24 & 360 & 51,840 \\
\underline{statues} ({armadillo}, {buddha}, {lucy}, {roman}, {thai})  & 24 & 360 & 43,200 \\
\underline{vehicles} ({bicycle}, {car}, {jeep}, {ship}, {spacehship}) & 24 & 360 & 43,200 \\
\underline{misc} ({diego}, {kettle}, {plants}, {teapot}, {skeleton}) & 24 & 360 & 43,200 \\
\hline
\end{tabularx}
\end{table}

% Each object in this subset has one 3D model. We provide images, depth maps, and surface normal maps for each object. Changes in lighting and shadows are some of the few noticeable visual cues for texture-less objects with otherwise homogeneous surfaces. However, these changes are not enough to indicate their 3D shape. We render these objects from multiple angles and under different lighting conditions to aid networks in learning illumination-independent features for object reconstruction. 

\Figure[ht][width=0.32\textwidth]{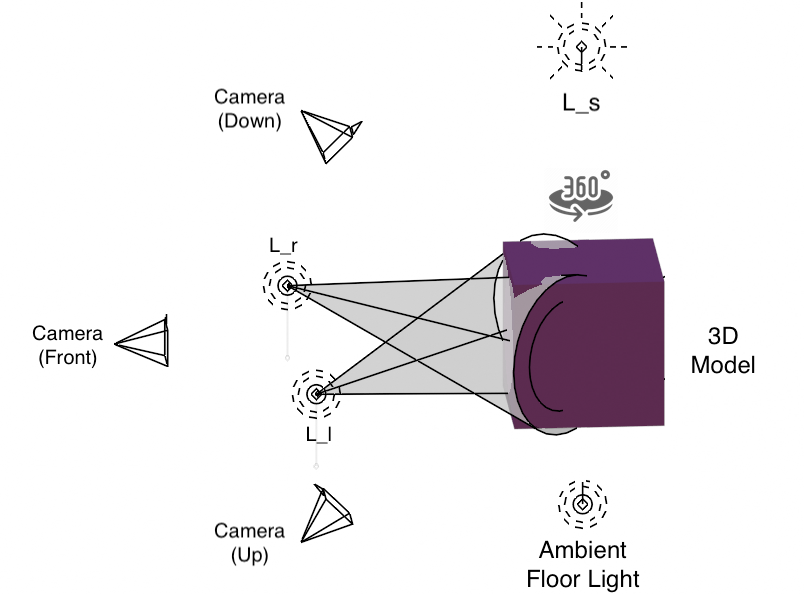}
{\textbf{Synthetic (A) Rendering Scene.} Multiple lights illuminate a rotating object. We render multiple views using three cameras.\label{fig:blender_scene}}

\begin{table}[ht]
\caption{\textbf{Synthetic (A) Rendering Configurations.} We use different combinations of lights, cameras, and colors to define 24 sequences.}
\label{tab:blender_config}
\begin{tabularx}{\linewidth}{|llX|}
\hline
& Configuration  & Description                     \\
\hline
Lights            & {sun}      & Only sunlight on.                        \\
                  & {left}     & Left lamp and sunlight on.               \\
                  & {right}   & Right lamp and sunlight on.              \\
                  & {all}     & Both lamps and sunlight on.              \\
\hline
Camera            & {down}    & Above the object, looking down.          \\
                  & {front}   & At object height, looking straight.      \\
                  & {up}      & Below the object, looking up.            \\
\hline
Color             & yes       & Rendered with a uniform random color. \\
                  & no        & Rendered in white. \\
\hline
\end{tabularx}
\end{table}

\noindent \textbf{Scene Setup.} \hspace{2pt} Inspired by the real-world scene used in~\cite{bednarik2018learning} for data collection, which comprises the object in front of a stationary camera and three lights in the room, we created a Blender scene with multiple lights and cameras around the object. This is illustrated in Figure~\ref{fig:blender_scene}. Each camera looks at the object from a different perspective, and several lights ensure the object is adequately illuminated with realistic shadows.

There are three cameras in the scene. The first camera is positioned parallel to the object in front, while the other two are placed at slightly elevated angles above and below the object. The front camera remains stationary, but the exact tilt angle of the other two cameras is randomly changed to ensure a variety of elevations. Additionally, the object is rotated by small azimuth steps around itself, providing comprehensive coverage from all sides.

Like natural scenes, which have multiple light sources affecting how a surface is perceived, we include various lights to simulate realistic shadows and illuminations. Two bright spotlights shine on the object from the front-left and front-right sides. These lights have a warm white color resembling the standard incandescent lamps with an RGB value of $(1.0, 0.945, 0.875)$. A blue-tinted cool sunlight with RGB value $(0.785, 0.883, 1.0)$ is added overhead, and a soft glowing white ambient light is placed on the floor to light up the bottom faces of the object. The sunlight has a power of 100 watts per square meter, while the ambient floor light has a power of 2 watts per square meter. The spotlight brightness varies based on the object's distance from the cameras.

\noindent \textbf{Configurations.} \hspace{2pt} The scene is rendered under different lighting configurations (Table~\ref{tab:blender_config}) with one or more lights on at a time. This generates a wide range of shadows and lighting effects in the dataset. It can help deep learning networks comprehend the impact of shadows on shape perception, resulting in networks that are robust to lighting variations.  We note that texture-less surfaces do not necessarily lack color. Rather, they lack distinctive textures on their surfaces. We expand the definition of texture-less surfaces in~\cite{bednarik2018learning} from grayscale to include surfaces with a uniform color. Consequently, we render all objects initially without color or texture, and then again with a diffuse material that has a uniformly random color.

\Figure[!ht][width=\linewidth]{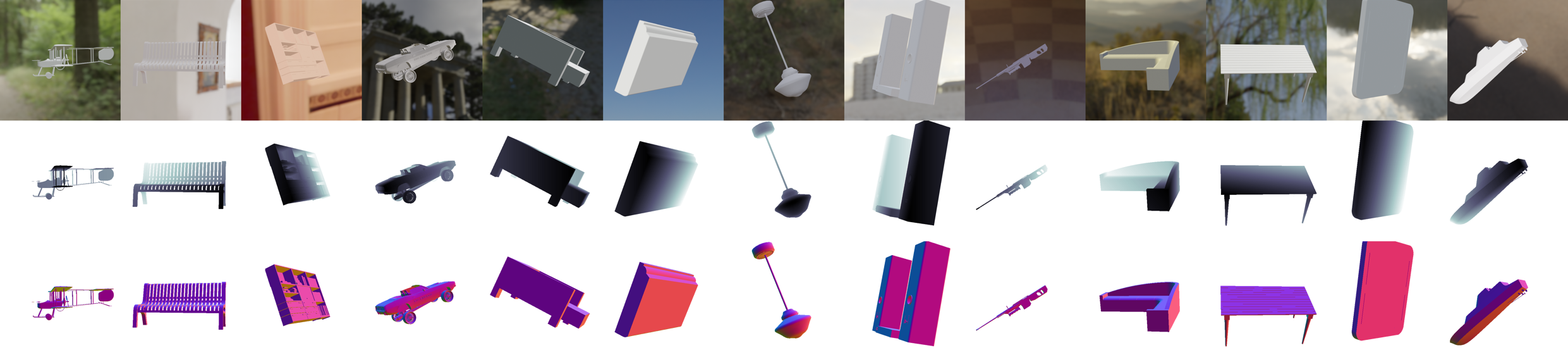}
{\textbf{Shape2.5D-Synthetic (B) Subset.} This set features 24 renderings lacking textures, comprising over 36,000 models of 13 principal ShapeNet objects set against real-world HDRI backgrounds. It includes RGB images, depth maps, and surface normals. The subset aims to emphasize variation in the shapes of the same object rather than rendering comprehensive perspectives on a restricted number of 3D models as found in the first set.\label{fig:data_shapenet}}

We use an automated script to generate data with Blender. The script creates a 3D scene with the required settings, including camera and lighting. It then enables rendering passes to capture RGB images and depth maps and sets up compositor nodes in Blender for post-processing rendered data. The 3D models are imported from Wavefront files located in the specified path. After importing, they are positioned at the origin with the appropriate scale and orientation, ensuring they are visible and upright inside the camera viewport. All textures are removed, leaving only the bare 3D model. We select one model at a time and rotate it through specified steps to render a sequence of views. This process is repeated for each enabled camera and lighting setup. For each view, an RGB image, a foreground mask, and a depth map are rendered. The rendered images have a resolution of 512x512 and a black background. Following~\cite{bednarik2018learning}, we differentiate depth maps to compute surface normals.

\begin{figure}[ht]
    \centering
    \subfloat[]{
        \includegraphics[scale=0.2]{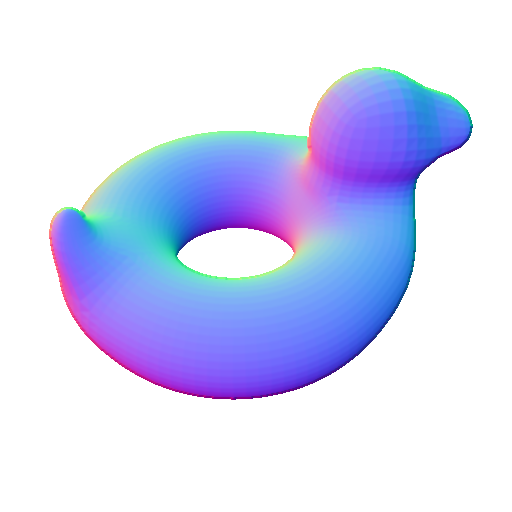}
    }
    \hfill
    \subfloat[]{
        \includegraphics[scale=0.2]{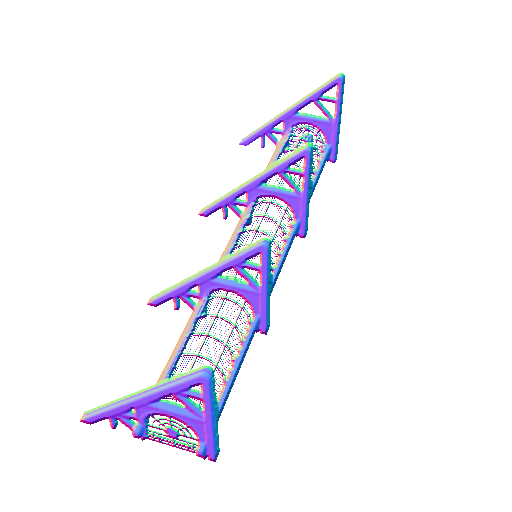}
    }
    \hfill
    \subfloat[]{
        \includegraphics[scale=0.2]{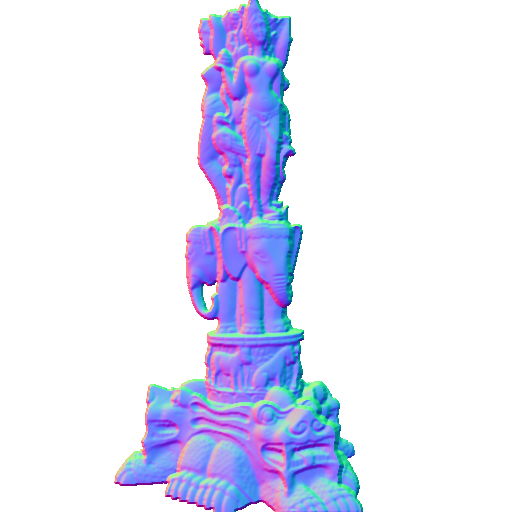}
    }
    \caption{\textbf{Depth Complexity:} (a) A simple rubber duck, (b) the detailed San Diego Convention Center, and (c) an ornate Thai statue.}
    \label{fig:objects_scale}
\end{figure}

\noindent \textbf{Depth Complexity.} \hspace{2pt} Our dataset comprises objects with varying degrees of realism regarding deformations and size variations. These objects range from a tiny rubber duck to several life-size statues and a model building (Figure~\ref{fig:objects_scale}).

\subsubsection{Synthetic (B) Subset}

\begin{figure}[ht]
    \centering
    \subfloat{
        \includegraphics[width=0.31\linewidth]{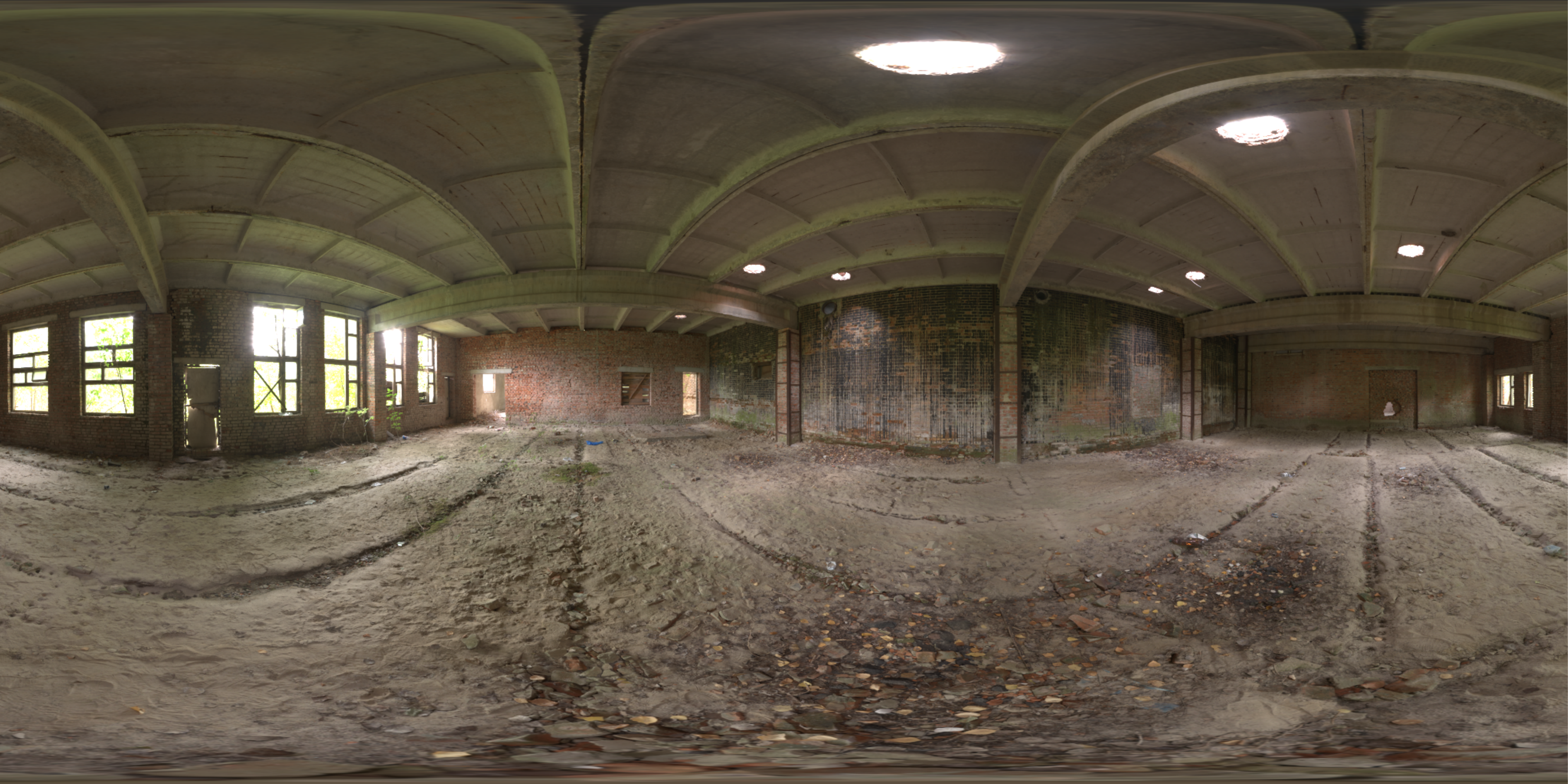}
    }
    \hfill
    \subfloat{
        \includegraphics[width=0.31\linewidth]{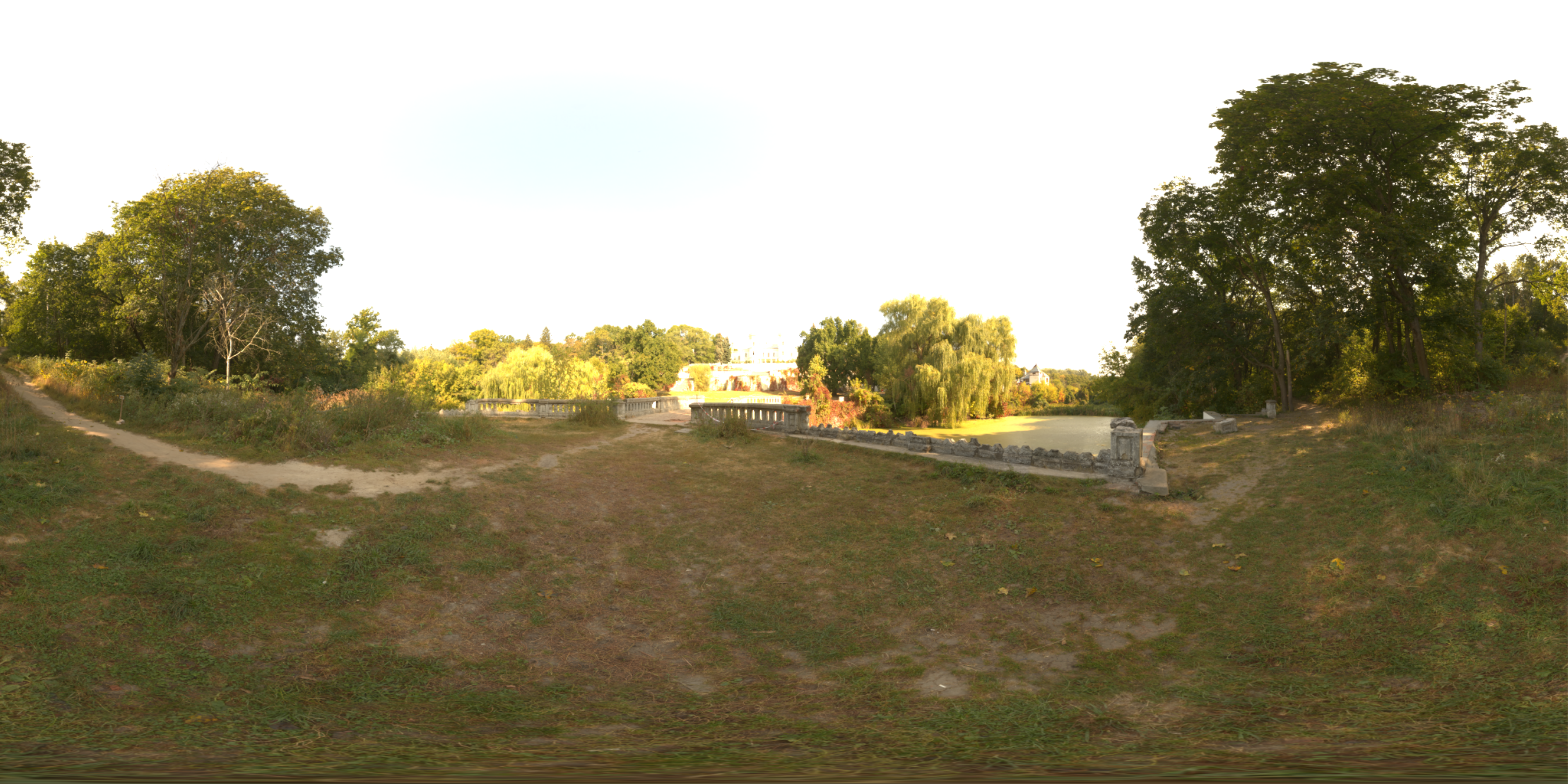}
    }
    \hfill
    \subfloat{
        \includegraphics[width=0.31\linewidth]{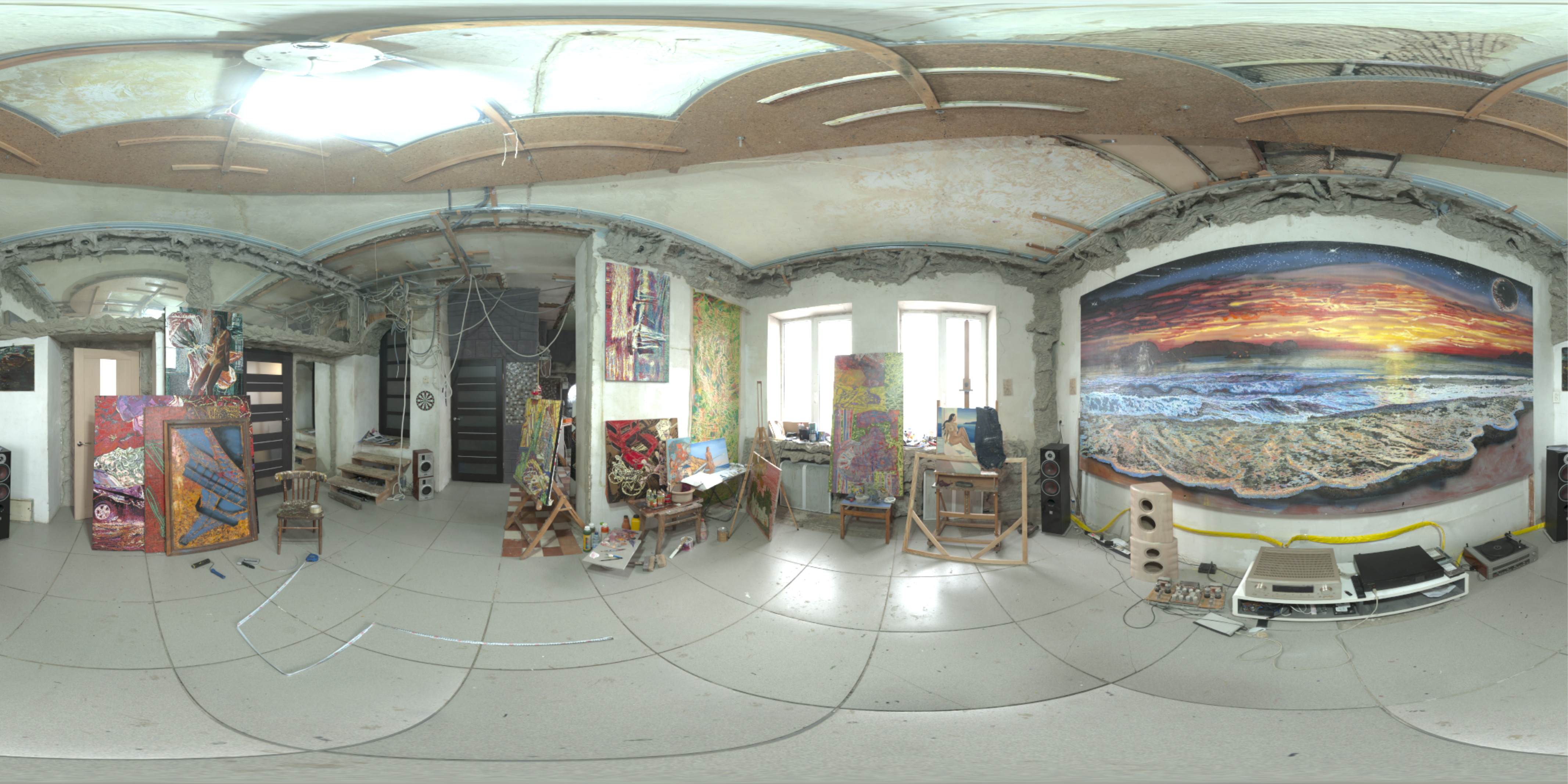}
    }
    \hfill
    \subfloat{
        \includegraphics[width=0.31\linewidth]{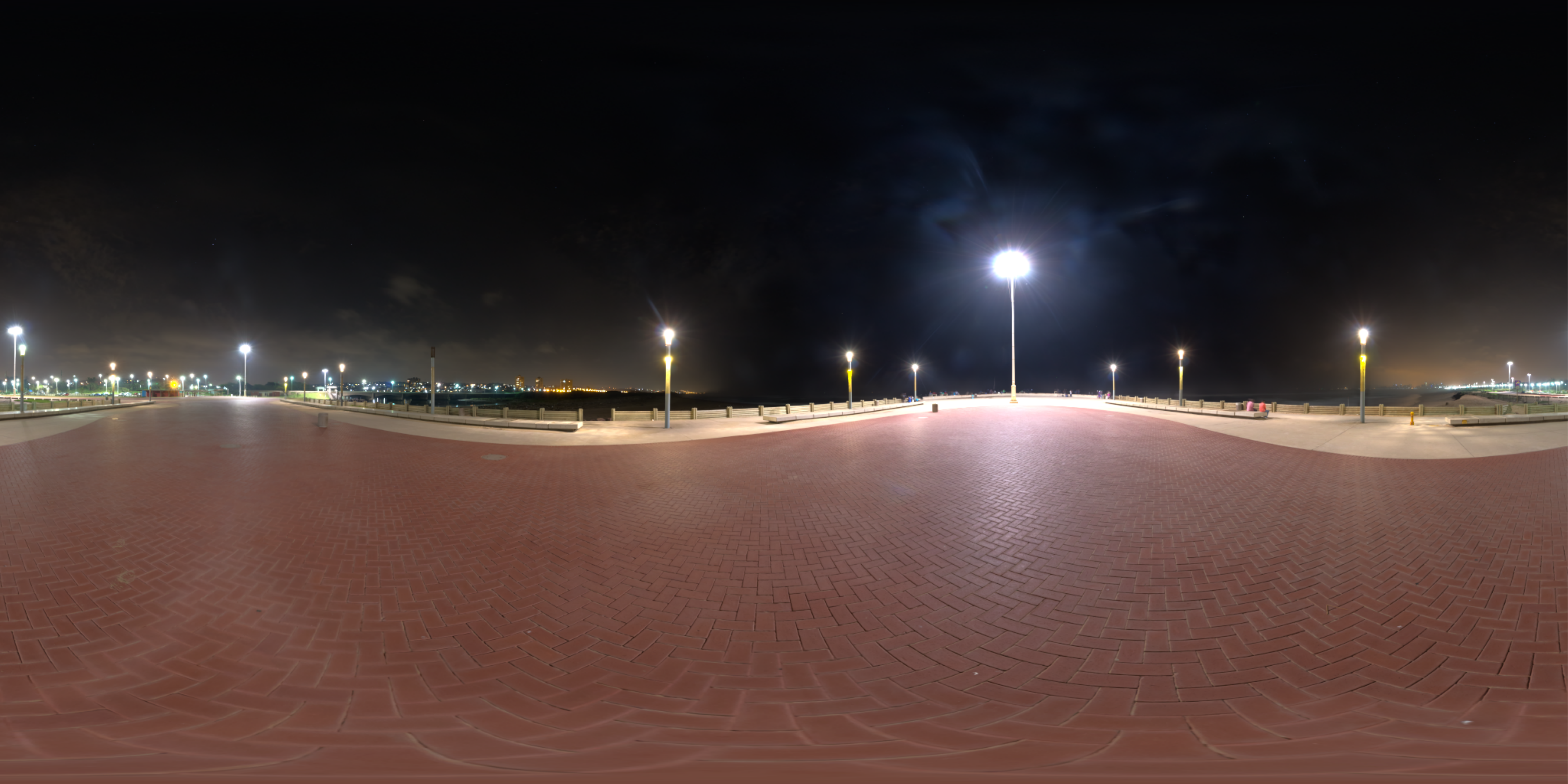}
    }
    \hfill
    \subfloat{
        \includegraphics[width=0.31\linewidth]{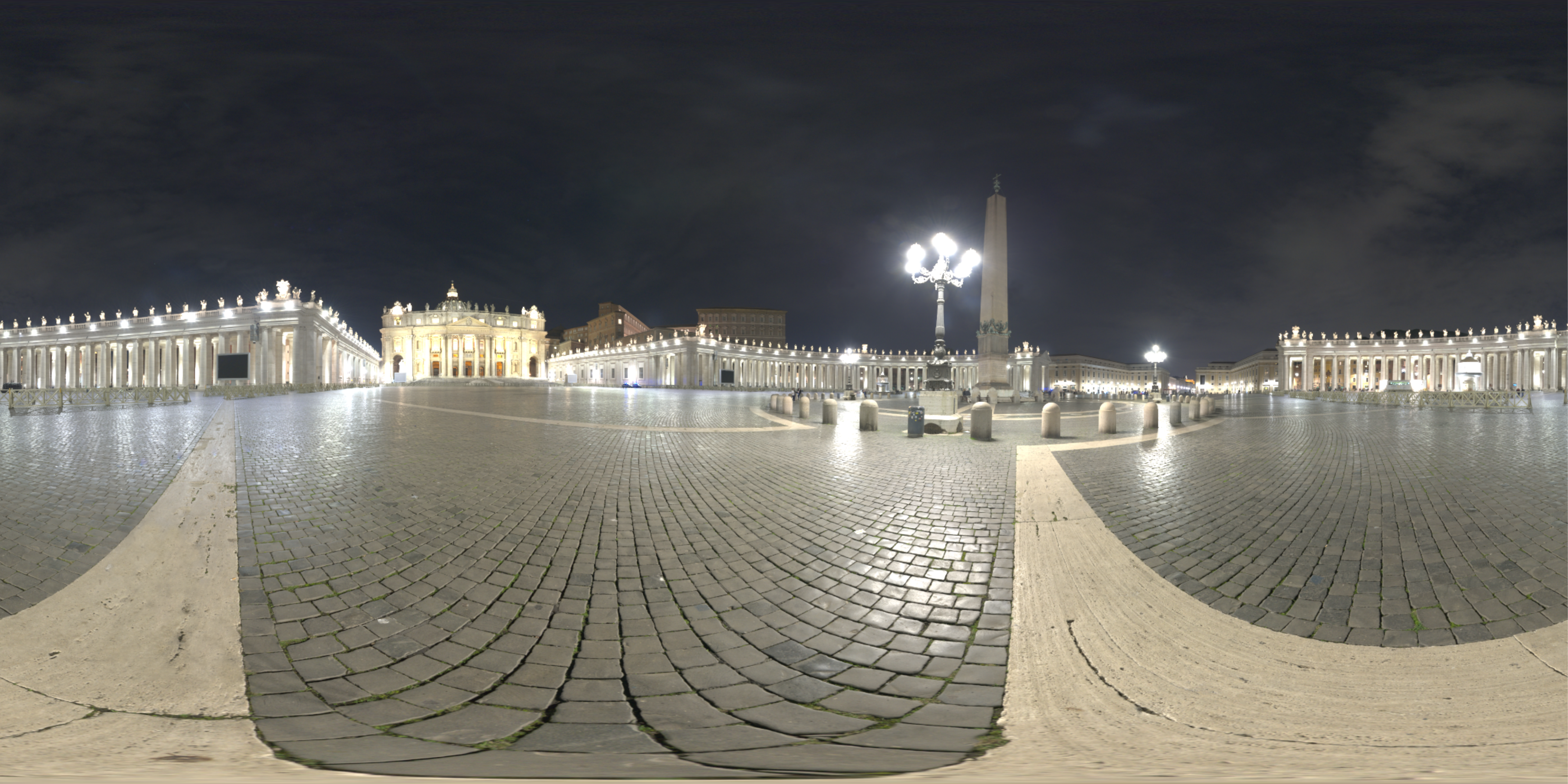}
    }
    \hfill
    \subfloat{
        \includegraphics[width=0.31\linewidth]{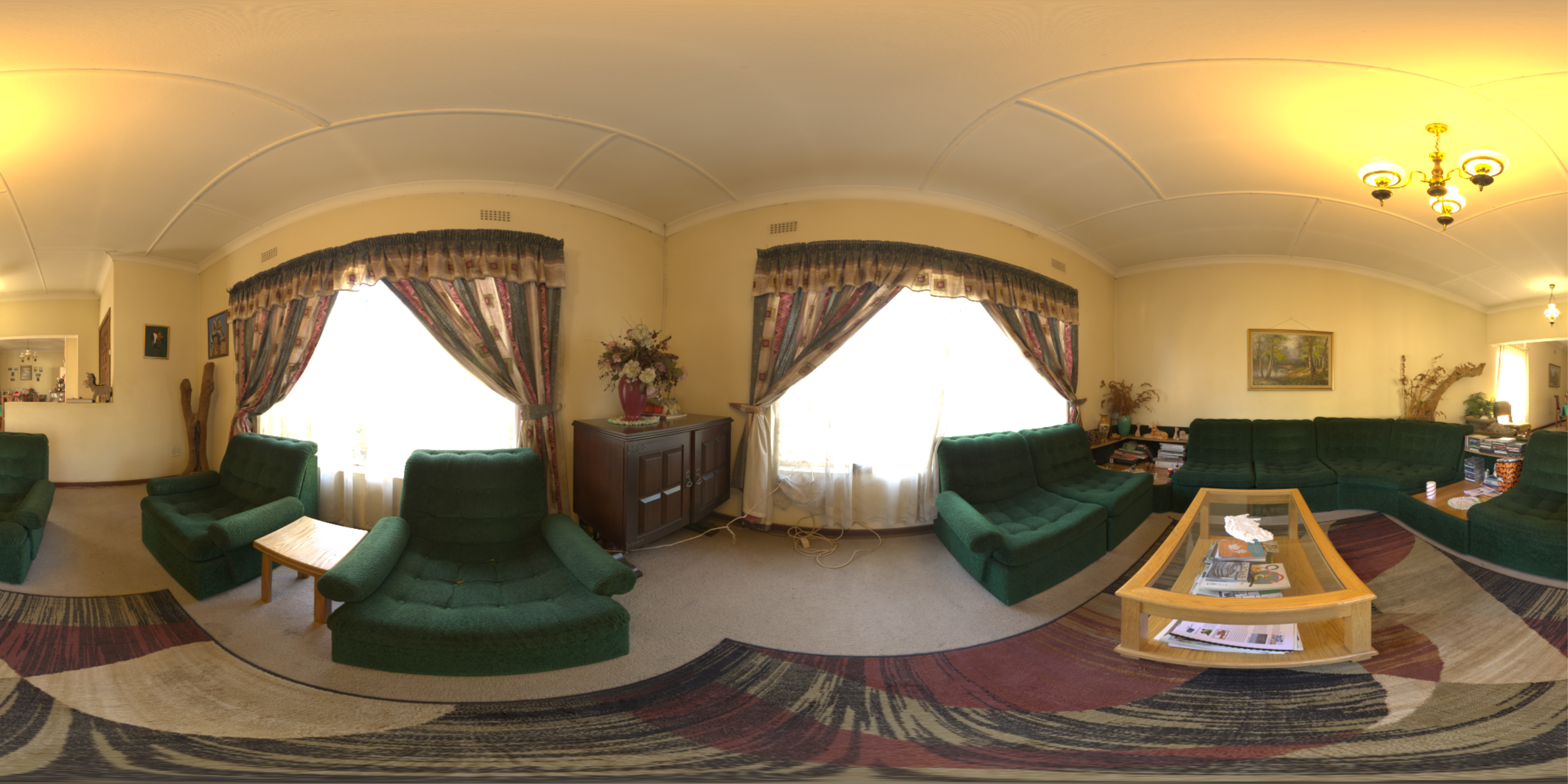}
    }
    \caption{\textbf{HDRI Environments.} We use 579 real-world backgrounds, including indoor and outdoor scenes during the day and night.}
    \label{fig:hdri_environments}
\end{figure}

This subset, shown in Figure~\ref{fig:data_shapenet}, comprises 13 objects selected from the Choy et al.~\cite{choy20163d} subset of the ShapeNet dataset~\cite{chang2015shapenet}. We render the same 13 object categories on 579 real-world HDRI environment maps (Figure~\ref{fig:hdri_environments}) in Blender, including both indoor and outdoor backgrounds. The lighting is computed directly from the environment.

\begin{figure}[ht]
    \centering
    \includegraphics[width=0.79\linewidth]{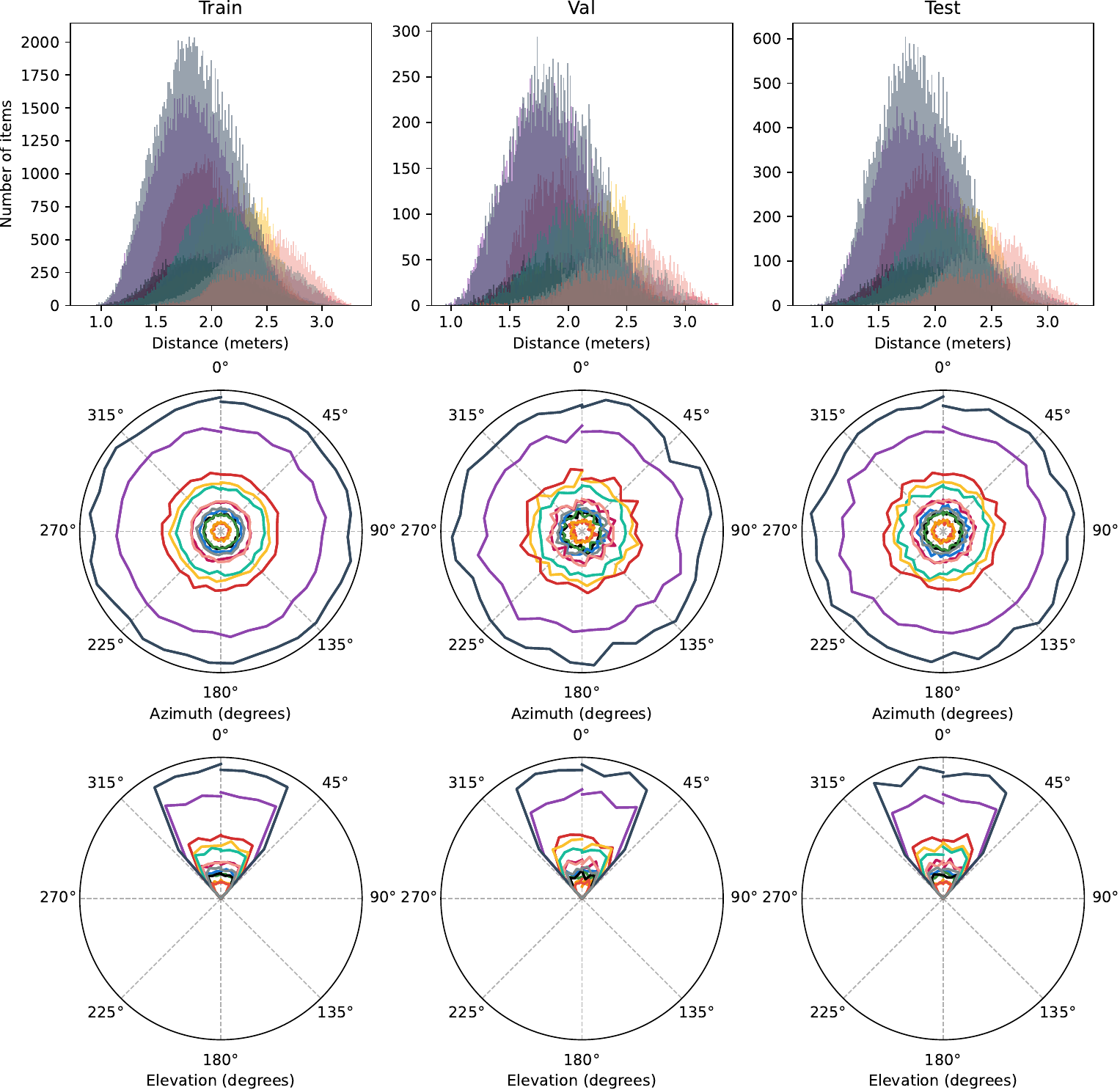}
    \includegraphics[width=0.2\linewidth]{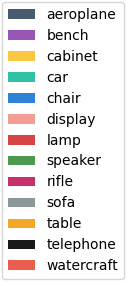}
    \caption{\textbf{Synthetic (B) Rendering Configurations.} We render over 39k 3D models from ShapeNet from different angles. The azimuth angles range from 0 to 360 degrees, while the elevation angles are between -45 and 45 degrees (i.e., 0-45 and 315-360). The in-plane rotation and camera field of view are always fixed at 0 and 25 degrees, respectively. The scale of the shapes is varied by adjusting the camera distance in the range of 1 and 3.5.\label{fig:metadata_textured}}
\end{figure}

Unlike the first set, where we rotate the object and uniformly sample views at different perspectives, we define a range of camera perspectives and sample 24 random views for this set to align with previous work on ShapeNet~\cite{choy20163d}. This is summarized in Figure~\ref{fig:metadata_textured}. We also provide depth maps, normal maps, and camera poses.

% \noindent \textbf{Rendering Environment} \hspace{2pt} We used Cycles, a ray-tracing engine, to render images on an Intel(R) Core(TM) i7-6700K CPU with a 4.00GHz processor. Each sample took around 30 seconds to generate. This time reduces to 5 seconds if a GPU rendering with CUDA is used. Our rendering script also supports the physically-based Eevee renderer.

\subsection{Real-World Dataset}
\label{sec:dataset_real}

To supplement the synthetic texture-less data, we also collected a small dataset of real-world low-texture objects containing 4,672 samples from six objects. This includes four deformable objects \textit{hoody}, \textit{shirt}, \textit{shorts}, and \textit{tshirt}, and two rigid objects \textit{chair} and \textit{lamp}. Unlike the dataset used by~\cite{bednarik2018learning}, which relied on controlled artificial lighting, we collected our data under natural daylight and multiple random artificial light sources. We explain the collection process in the following section.~Table~\ref{tab:data_real} provides a summary of the objects in our real-world dataset.

\subsubsection{Collection Process}

We recreated the setup depicted in~Figure~\ref{fig:blender_scene} in a real-world environment. The Kinect was mounted on a tripod with adjustable height, facing away from a large window that allowed natural light into the room. The room was illuminated by a combination of natural light from the window and four different fluorescent light bulbs. The objects were placed at distances ranging from 0.5m to 1.25m from the camera. As the accuracy of depth values obtained through Kinect can be affected by temperature~\cite{wasenmuller2016comparison}, we allowed the camera to run for 30 minutes to reach a stable temperature before starting data capture each time. We varied the camera height and viewing angle to obtain the \textit{up}, \textit{down}, and \textit{front} camera positions. The light bulbs were switched on and off randomly to provide different lighting across sequences. Each sequence had at least two lights to create complex shadows.

For data collection, the clothing pieces were worn by an individual who moved randomly in front of a camera. Meanwhile, other objects were positioned on a stationary surface and manually rotated in front of the camera. The camera captured synchronized RGB images and depth maps. In the post-processing stage, described in the next section, a segmentation algorithm selected only the object of interest and removed the person and any background. The surface normals were then computed by smoothing the depth maps using a $5 \times 5$ Gaussian filter and differentiating them.

\Figure[ht]()[width=0.45\textwidth]{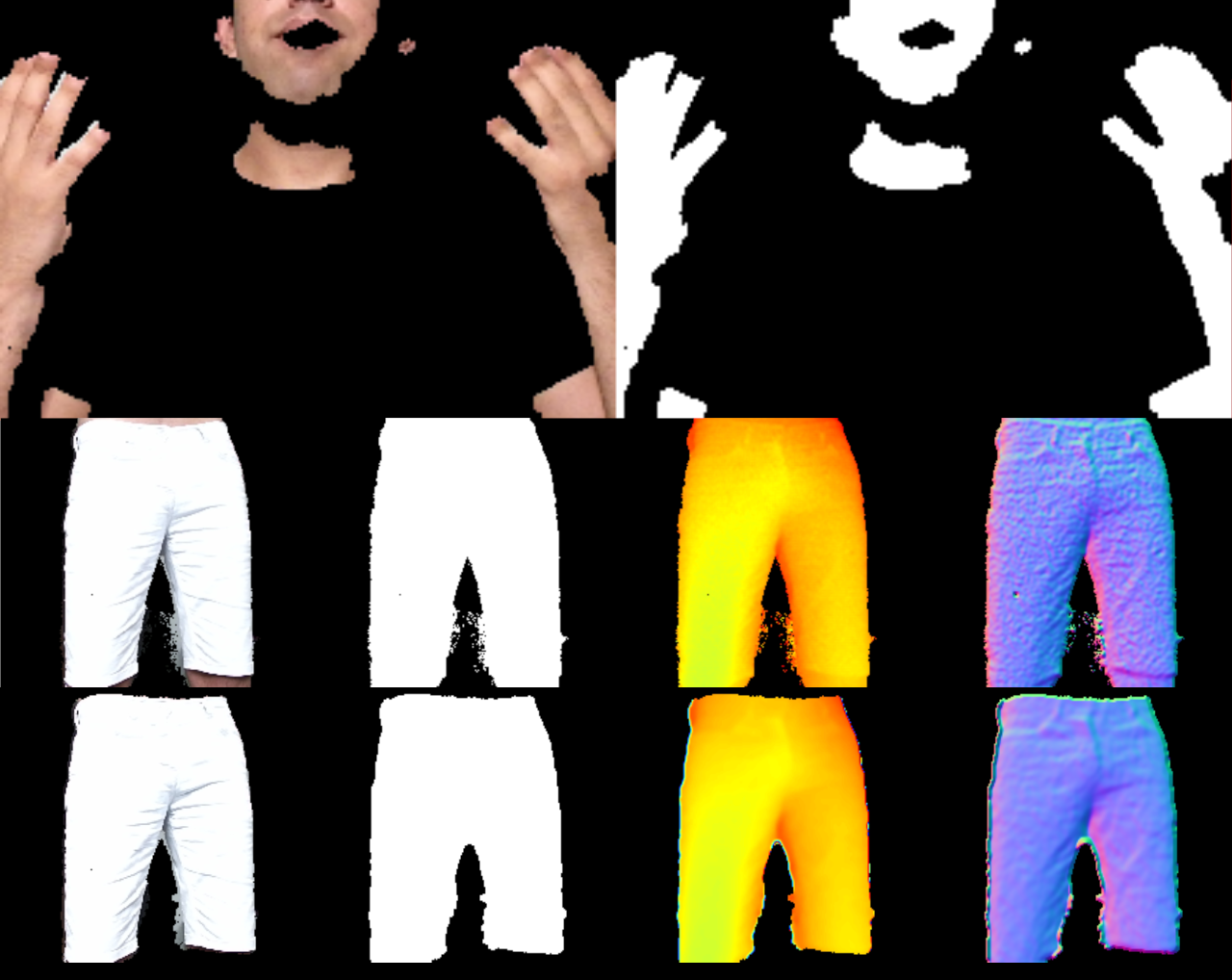}
{The top row of the figure shows the results of our skin-detection algorithm that removes the person wearing the clothes from the images. The middle row shows the raw output from Kinect with a lot of noise and a hole in the depth and normal maps (right leg). The bottom row shows the output after post-processing.\label{fig:postproc}}

\begin{table}[ht]
\caption{Summary of objects in the dataset of real objects.}
\label{tab:data_real}
\setlength{\tabcolsep}{3pt}
\begin{tabularx}{\linewidth}{|p{1.5cm}|YYYYYY|}
\hline
Object    & {hoody} & {shirt} & {shorts} & {tshirt} & {chair} & {lamp} \\
\hline
Sequences &    3 &    4 &    1 &    9 &    2 &    1 \\
Samples   &  508 &  671 &  387 & 1545 & 1201 &  360 \\
\hline
\end{tabularx}
\end{table}

\subsubsection{Post-Processing}

After capturing raw data with the Kinect camera, we perform several post-processing steps to prepare it for use. These steps are summarized in~Figure~\ref{fig:postproc}.

\noindent \textbf{Hole filling. } \hspace{2pt} The Kinect uses phase-shifted infrared light to compute depth, but it can sometimes miss surfaces that absorb infrared waves, are translucent, or very shiny. This results in holes in the depth map. To fix this, we used OpenCV's implementation of the Navier-Stokes-based method~\cite{bertalmio2001navier} of inpainting with a radius of 7, which interpolates the missing parts.

\noindent \textbf{Background segmentation. } \hspace{2pt} To identify the body of a person wearing clothes, a naive skin-color detection algorithm initialized with an estimated skin color of the person was used. For objects placed on a flat surface, the surface was covered with a skin-colored cloth. In the captured image, any skin-colored pixels or those more than 1.25 meters away from the camera were considered background. We removed the background and filtered out small contours as noise that escaped skin detection. The largest contour in the scene was identified as the foreground, which contained the object of interest. Finally, we performed a morphological closing operation to fill any small holes in the foreground mask. We have included a binary mask of the foreground for each sample in the dataset.

\noindent \textbf{Normalization. } \hspace{2pt} The normal vectors were transformed into unit-length vectors with values ranging from -1 to 1. This was achieved by dividing each vector by its magnitude. Additionally, all depth values in the dataset were normalized between 0 and 1. This was done to ensure that the dataset remains invariant to the arbitrary choice of camera distance from the object. The background pixels were assigned a depth value of 1, while the foreground pixels had values ranging between 0 and 0.99.

\section{Experiments}
\label{sec:benchmarks}

To exhibit the usability of our dataset, we evaluated it on a pre-existing neural network to estimate the depth and surface normals of texture-less objects.

\subsection{Depth and Normal Estimation Benchmarks}

\noindent \textbf{Baseline.} \hspace{2pt} Bednarik et al.~\cite{bednarik2018learning} showed that their encoder-decoder network is a good starting point for reconstructing depth and normal maps from a single image of deformable, texture-less surfaces. It is the closest work to our task, for which we created a new dataset. Inspired by their results, we used their network as a baseline for our benchmarks by removing the mesh decoder. However, unlike~\cite{bednarik2018learning}, we trained the depth and normal decoders jointly without using early stopping. We kept all other hyperparameters the same and added an edge consistency component to the original depth and normal loss functions of~\cite{bednarik2018learning} by computing the MSE loss for the gradients of the depth and normal maps. In all experiments, we trained the network for 50 epochs.

\noindent \textbf{Evaluation Metrics.} \hspace{2pt} The \textit{depth error metric}, denoted $E_D$, measures the absolute difference between the predicted and actual depth values, considering only the foreground pixels. Since the depth values are normalized between 0 and 1, the metric does not represent real-world units but rather a percentage. We represent the mean absolute difference in depth values as a percentage multiplied by 100. The \textit{normal error metric}, denoted as $E_N$, is the mean angular distance in degrees between the surface normals of the predicted and actual values. Smaller values of $E_D$ and $E_N$ indicate better results. Furthermore, we report the percentage of surface normals with errors of less than 10$^{\circ}$, 20$^{\circ}$, and 30$^{\circ}$, respectively. Higher percentages indicate a better quality of the normal vectors. Only foreground pixels are used to calculate all metrics, as background pixels are not labeled.

\noindent \textbf{Data Splits.} \hspace{2pt} We split the 24 configurations of the Synthetic (A) data set by camera and lighting conditions as summarized in Table~\ref{tab:data_synthetic}. For Synthetic (B), standard ShapeNet splits are used for training, validation, and testing. The real-world dataset is only used for testing.

\begin{table}[ht]
\caption{\textbf{Proposed splits for Synthetic (A).} We split the dataset into 16 training sequences, 4 validation sequences, and 4 test sequences.}
\label{tab:data_synthetic}
\begin{tabularx}{\linewidth}{|X|p{1.5cm}|p{1.5cm}|}
\hline
Train & Val & Test \\
\hline
$S_\mathrm{left-down}$, $S_\mathrm{left-front}$, $S_\mathrm{left-up}$, $S_\mathrm{right-down}$, $S_\mathrm{right-front}$, $S_\mathrm{right-up}$, $S_\mathrm{sun-down}$, $S_\mathrm{sun-front}$ 

& $S_\mathrm{all-down}$, $S_\mathrm{all-front}$ 

& $S_\mathrm{all-up}$, $S_\mathrm{sun-up}$ \\

% \hline
% B & $L_{a_{down}}$, $L_{a_{up}}$, $L_{a_{front}}$ $(0^\circ$, $45^\circ$, $90^\circ, 180^\circ$, $225^\circ$, $270^\circ)$ & $L_{a_{down}}$, $L_{a_{up}}$, $L_{a_{front}}$ $(135^\circ)$ & $L_{a_{down}}$, $L_{a_{up}}$, $L_{a_{front}}$ $(315^\circ)$ \\
\hline
\end{tabularx}
\end{table}

\subsubsection{Intra-Category Benchmark}

Intra-category experiments on Synthetic (A) involve training and testing inside the same category to evaluate the network's generalization to new lighting conditions for previously encountered views.

\begin{table}[ht]
\caption{\textbf{Intra-Category Benchmark.} The clothing and furniture categories show the best results with more than 93.88\% and 91.17\% of the predicted normals, respectively, having smaller than 30$^{\circ}$ angular difference from the groundtruth. This shows that our dataset allows for a good inter-class generalization ability.}
\label{tab:results_categories}
\setlength{\tabcolsep}{3pt}
\begin{tabularx}{\linewidth}{|p{1.5cm}|c|cYYY|}
\hline
Dataset & \textbf{$E_D$} & $E_N$ & $<$10$^{\circ}$ & $<$20$^{\circ}$ & $<$30$^{\circ}$ \\
\hline
animals	 &  5.02$\pm$ 5.02 & 14.19$\pm$10.75 & 63.83 & 81.84 & 88.56 \\
clothing	 &  2.61$\pm$ 1.66 & 10.59$\pm$ 2.64 & 67.69 & 87.47 & 93.88 \\
furniture	 &  3.79$\pm$ 5.73 & 11.67$\pm$ 8.13 & 69.33 & 84.82 & 91.17 \\
misc	     &  2.99$\pm$ 1.71 & 16.98$\pm$10.25 & 58.25 & 73.38 & 81.63 \\
statues	     &  5.83$\pm$ 7.06 & 16.58$\pm$10.02 & 50.76 & 75.31 & 86.00 \\
vehicles	 &  4.08$\pm$ 5.60 & 16.31$\pm$ 8.61 & 55.41 & 73.88 & 83.62 \\
\hline
\end{tabularx}
\end{table}

Table~\ref{tab:results_categories} shows the performance of the baseline network trained on objects in each of the six main categories and evaluated on a subset of the same objects in unseen lighting conditions. The results demonstrate that the network successfully learns the shape of the texture-less objects under illumination changes reasonably well.

\subsubsection{Inter-Category Benchmark}

In inter-category experiments on Synthetic (A), models are trained on one category and then assessed on the others to examine the network's ability to generalize across different shape categories and objects.

\begin{table}[ht]
\caption{\textbf{Inter-Category Benchmark.} The degree of generalization to new categories is less than the intra-category experiments. The network learns strong shape priors that do not generalize well to very different geometries.}
\label{tab:results_objects}
\setlength{\tabcolsep}{3pt}
\begin{tabularx}{\linewidth}{|Y|Y|c|cccc|}
\hline
Train & Test & $E_D$ & $E_N$ & $<$10$^{\circ}$ & $<$20$^{\circ}$ & $<$30$^{\circ}$ \\
\hline
\multirow{5}{*}{{animals}}
&	clothing       & 20.76$\pm$ 6.34 & 26.71$\pm$ 7.79 & 15.93 & 50.65 & 72.43 \\
&	furniture      & 21.28$\pm$ 6.91 & \textbf{26.09$\pm$ 8.99} & \textbf{22.58} & \textbf{53.13} & \textbf{73.27} \\
&	misc           & \textbf{20.72$\pm$ 9.07} & 43.63$\pm$19.47 & 14.37 & 36.85 & 51.56 \\
&	statues        & 26.37$\pm$ 7.16 & 41.37$\pm$14.86 & 11.55 & 36.26 & 56.71 \\
&	vehicles       & 20.83$\pm$ 4.69 & 53.61$\pm$24.12 &  9.48 & 27.09 & 41.98 \\
\hline
\multirow{5}{*}{{clothing}}
&{animals}   & \textbf{19.35$\pm$ 4.78} & \textbf{25.18$\pm$ 4.31} & \textbf{18.13} & \textbf{52.36} & \textbf{72.06} \\
&{furniture} & 20.50$\pm$ 4.54 & 29.28$\pm$ 7.18 & 13.65 & 39.79 & 64.97 \\
&{misc}      & 20.61$\pm$ 6.16 & 39.39$\pm$10.05 & 11.43 & 30.97 & 48.97 \\
&{statues}   & 19.85$\pm$ 5.43 & 29.08$\pm$ 6.70 & 17.48 & 44.56 & 64.9  \\
&{vehicles}  & 22.91$\pm$ 5.70 & 38.90$\pm$10.10 &  9.30 & 28.19 & 47.02 \\
\hline
\multirow{5}{*}{{furniture}}
&{animals}   & \textbf{22.40$\pm$ 6.02} & \textbf{31.14$\pm$ 7.47} & \textbf{14.04} & \textbf{40.22} & \textbf{61.73} \\
&{clothing}  & 24.84$\pm$ 5.29 & 35.53$\pm$ 8.82 &  8.92 & 30.96 & 54.44 \\
&{misc}      & 23.07$\pm$ 8.65 & 43.79$\pm$14.35 & 10.75 & 29.13 & 45.83 \\
&{statues}   & 29.57$\pm$ 8.29 & 41.20$\pm$11.80 &  8.42 & 28.00 & 49.57 \\
&{vehicles}  & 24.13$\pm$ 8.26 & 49.17$\pm$18.60 & 10.89 & 27.22 & 41.55 \\
\hline
\multirow{5}{*}{{misc}}
&{animals}   & 21.55$\pm$ 5.54 & 38.49$\pm$ 7.42 & 11.24 & 31.85 & 50.48 \\
&{clothing}  & 25.76$\pm$ 5.89 & 35.46$\pm$ 6.01 &  8.36 & 30.40 & 52.85 \\
&{furniture} & 23.36$\pm$ 5.73 & 36.95$\pm$ 8.57 & 14.38 & 35.58 & 53.89 \\
&{statues}   & \textbf{19.77$\pm$ 6.41} & \textbf{33.85$\pm$ 9.95} & \textbf{18.80} & \textbf{38.04} & \textbf{54.93} \\
&{vehicles}  & 21.15$\pm$ 6.82 & 48.60$\pm$14.74 &  8.47 & 24.17 & 39.58 \\
\hline
\multirow{5}{*}{{statues}}
&{animals}   & \textbf{20.85$\pm$ 5.15} & \textbf{38.34$\pm$ 7.54} & \textbf{12.14} & \textbf{35.67} & \textbf{55.39} \\
&{clothing}  & 22.66$\pm$ 5.14 & 40.87$\pm$ 9.28 &  7.93 & 28.52 & 51.64 \\
&{furniture} & 23.78$\pm$ 5.21 & 48.09$\pm$11.43 &  9.72 & 30.26 & 47.85 \\
&{misc}      & 22.70$\pm$ 6.07 & 49.31$\pm$10.18 &  8.62 & 25.32 & 40.49 \\
&{vehicles}  & 23.85$\pm$ 5.48 & 50.93$\pm$12.51 &  7.37 & 23.95 & 40.36 \\
\hline
\multirow{5}{*}{{vehicles}}
&{animals}   & 23.24$\pm$ 6.67 & \textbf{29.28$\pm$ 4.95} & 15.57 & \textbf{44.04} & \textbf{65.01} \\
&{clothing}  & 23.29$\pm$ 7.87 & 32.59$\pm$ 6.79 & 15.90 & 40.90 & 59.69 \\
&{furniture} & \textbf{22.08$\pm$ 5.87} & 32.92$\pm$ 6.58 & \textbf{16.21} & 41.33 & 60.14 \\
&{misc}      & 23.90$\pm$ 7.17 & 41.13$\pm$12.24 & 13.61 & 34.77 & 49.95 \\
&{statues}   & 29.83$\pm$ 9.69 & 31.36$\pm$ 6.63 & 13.80 & 39.77 & 61.51 \\
\hline
\end{tabularx}
\end{table}

\Figure[ht]()[width=\textwidth]{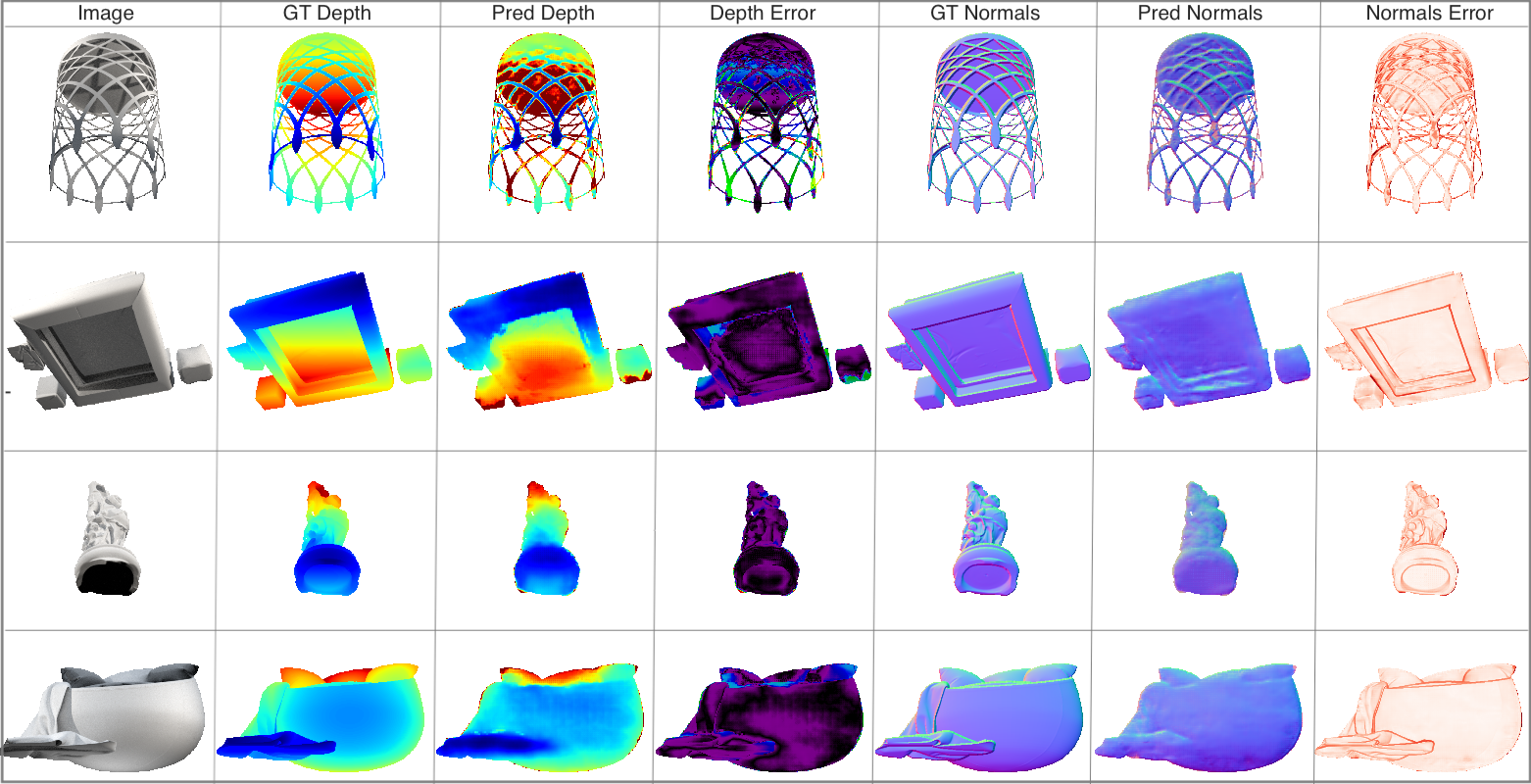}
{\textbf{Qualitative Results of Depth and Normal Estimation Benchmarks on Synthetic (A).} Visualization of qualitative errors in random test samples. \label{fig:qualitative_errors}}

Objects in a single category sometimes share similar geometric structures. For example, all statues have humanoid shapes with faces, torsos, arms, legs, and many small but gradual surface orientation changes. On the other hand, furniture mainly consists of plain surfaces with sharp corners and planes, often bending at right angles.~Table~\ref{tab:results_objects} lists the results of inter-category experiments with networks trained on one category of objects and evaluated on objects in all other categories. These experiments show the ability of the network to learn geometry-independent features and generalize to new types of texture-less objects.

Figure~\ref{fig:qualitative_errors} visualizes three random furniture items and a statue predicted by the network trained on the synthetic furniture category, showing the strongest errors near the edges.

\subsubsection{ShapeNet Benchmark}

We also benchmark the ShapeNet renders in Synthetic (B) for the depth and normal estimation tasks. In this experiment, renders of only the first 200 3D models for each of the 13 ShapeNet objects are used. Table~\ref{tab:results_shapenet} shows the results. During training, the network sees 18 of 24 renders for each model. Results are evaluated on six unseen views for each 3D model and reported individually for each object. These results are significantly better than the main categories because of a much higher number of 3D models.

\begin{table}[ht]
\caption{\textbf{Synthetic (B) Depth and Normal Estimation Benchmark.} Performance improves when more shapes are seen during training, showing the network can learn shape representations from our texture-less renders.}
\label{tab:results_shapenet}
\setlength{\tabcolsep}{3pt}
\begin{tabularx}{\linewidth}{|p{1.5cm}|c|cYYY|}
\hline
Object & $E_D$ & $E_N$ & $<$10$^{\circ}$ & $<$20$^{\circ}$ & $<$30$^{\circ}$ \\
\hline
{plane}     & 17.59±7.73	& 35.35±19.50	& 24.56	& 41.64	& 52.46 \\
{bench}     & 13.05±4.63	& 25.60± 9.13	& 34.73	& 53.43	& 65.27 \\
{cabinet}   & \textbf{8.23±4.01}	& \textbf{12.60±6.26}	& \textbf{76.97}	& \textbf{85.01}	& \textbf{87.89} \\
{car}       &  8.26±3.84	& 22.14± 9.87	& 39.91	& 62.05	& 74.28 \\
{chair}     & 14.75±5.36	& 24.08±13.37	& 39.51	& 59.29	& 70.58 \\
{display}   & 13.13±6.07	& 22.53±12.51	& 50.91	& 67.26	& 75.14 \\
{lamp}      & 16.71±9.59	& 26.48± 9.97	& 25.13	& 47.56	& 64.55 \\
{speaker}   &  9.76±5.61	& 16.03±10.90	& 63.61	& 76.50	& 83.26 \\
{rifle}     & 17.71±7.94	& 33.97±16.57	& 21.60	& 40.50	& 53.67 \\ 
{sofa}      &  9.25±3.57	& 14.00± 6.71	& 62.51	& 78.66	& 86.25 \\
{table}     & 12.02±4.78	& 18.99±13.59	& 57.85	& 68.77	& 75.78 \\
{phone}     & 13.90±8.78	& 24.55±14.17	& 50.08	& 66.60	& 74.90 \\
{watercraft}& 15.24±8.13	& 28.06±15.60	& 32.85	& 50.63	& 62.63 \\
\hline
\end{tabularx}
\end{table}

\subsubsection{Real-World Benchmark}

We evaluate the baseline network trained on the 'clothing' subset of our synthetic dataset on this real-world data and report the results in~Table~\ref{tab:results_our_real}. The complete real dataset is used as the test set.~Figure~\ref{fig:qualitative_errors_real_our} shows the qualitative results of these experiments on a random sample from the dataset.

\begin{table}[ht]
\caption{\textbf{Shape2.5D-Real Benchmark.} Results of reconstructing depth and normal vectors of our real dataset using the baseline network trained on our synthetic clothing dataset.}
\label{tab:results_our_real}
\setlength{\tabcolsep}{3pt}
\begin{tabularx}{\linewidth}{|p{1.5cm}|c|cYYY|}
\hline
Object & $E_D$ & $E_N$ & $<$10$^{\circ}$ & $<$20$^{\circ}$ & $<$30$^{\circ}$ \\
\hline
chair   & 29.62$\pm$8.89 & 43.68$\pm$8.54 & 10.15 & 31.39 & 49.84 \\
hoody   & 36.60$\pm$8.54 & 36.55$\pm$3.68 &  9.38 & 31.37 & 55.20 \\
lamp    & 23.30$\pm$6.27 & 74.26$\pm$7.07 &  3.39 & 11.85 & 21.88 \\
shirt   & 36.99$\pm$7.85 & 36.24$\pm$4.41 &  8.88 & 31.54 & 56.49 \\
shorts  & 42.35$\pm$7.12 & 26.78$\pm$3.01 & 12.68 & 42.35 & 73.03 \\
tshirt  & 36.12$\pm$9.22 & 35.45$\pm$3.38 & 10.20 & 34.77 & 59.92 \\
\hline
\end{tabularx}
\end{table}

\Figure[ht]()[width=0.45\textwidth]{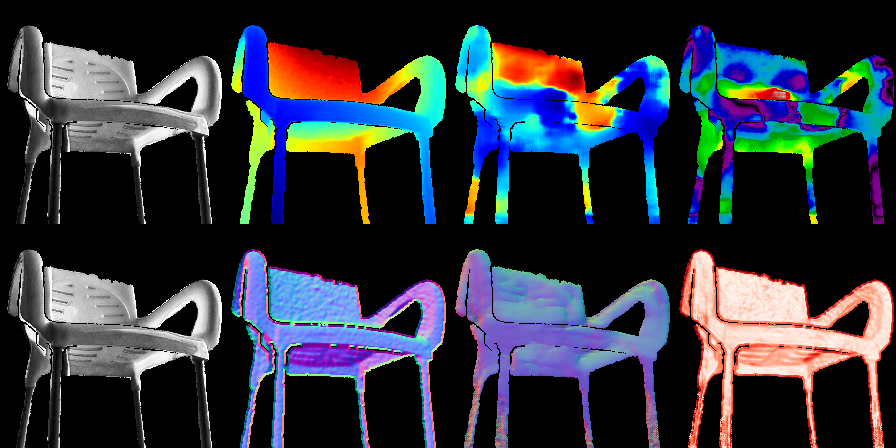}
{\textbf{Qualitative Results of Shape2.5D-Real Benchmark.} Visualization of the qualitative errors on a random sample from our real-world test data.\label{fig:qualitative_errors_real_our}}

To further demonstrate the generalization capability of the network to real-world data, we choose the real texture-less dataset from Bednarik et al.~\cite{bednarik2018learning}. This dataset has five items, and~\cite{bednarik2018learning} defines an experiment where the $cloth$ object is used to train the network, and 100 samples from each object are used for testing. We train the network on the real images of $cloth$ object from~\cite{bednarik2018learning}. In~Table~\ref{tab:results_real}, this network's performance for the surface normals map reconstruction task is compared with the network trained on the synthetic clothing category. Results for the {shapenet} category are provided in the supplemental material, where results of experiments on another real-world dataset that we created are also reported.

\begin{table}[ht]
\caption{\textbf{Bednarik-Real Benchmark.} Comparison of normal map reconstruction between a network trained on our synthetic dataset (\textbf{S}) and the same network trained on real $cloth$ data from~\cite{bednarik2018learning} (\textbf{R}). When trained on our synthetic data, the same network gives better surface normals for all four real objects other than the $cloth$ object used to train the real network, where our results are comparable.}
\label{tab:results_real}
\setlength{\tabcolsep}{3pt}
\begin{tabularx}{\linewidth}{|p{1.5cm}c|cYYY|}
\hline
Test & Train & $E_N$ & $<$10$^{\circ}$ & $<$20$^{\circ}$ & $<$30$^{\circ}$ \\
\hline
cloth   & \textbf{S} & 37.58 $\pm$  5.95 &  6.46 & 22.46 & 42.92 \\
        & \textbf{R} & \textbf{33.96 $\pm$  6.47} & \textbf{11.31} & \textbf{32.95} & \textbf{53.40} \\
\hline
hoody   & \textbf{S} & \textbf{36.96 $\pm$  2.31} &  \textbf{6.58} & \textbf{23.95} & \textbf{45.74} \\
        & \textbf{R} & 40.94 $\pm$  2.85 &  5.29 & 18.95 & 36.70 \\
\hline
paper   & \textbf{S} & \textbf{43.87 $\pm$  6.92} &  \textbf{4.28} & \textbf{16.08} & \textbf{32.24} \\
        & \textbf{R} & 45.91 $\pm$  7.32 &  4.02 & 15.19 & 30.54 \\
\hline
sweater & \textbf{S} & \textbf{40.66 $\pm$  2.66} &  \textbf{4.73} & \textbf{17.90} & \textbf{36.51} \\
        & \textbf{R} & 47.85 $\pm$  3.67 &  3.26 & 12.60 & 26.45 \\
\hline
tshirt  & \textbf{S} & \textbf{35.89 $\pm$  4.27} &  \textbf{6.85} & \textbf{24.68} & \textbf{46.71} \\
        & \textbf{R} & 42.43 $\pm$  6.24 &  4.56 & 17.15 & 34.34 \\
\hline
\end{tabularx}
\end{table}

\Figure[ht]()[width=0.45\textwidth]{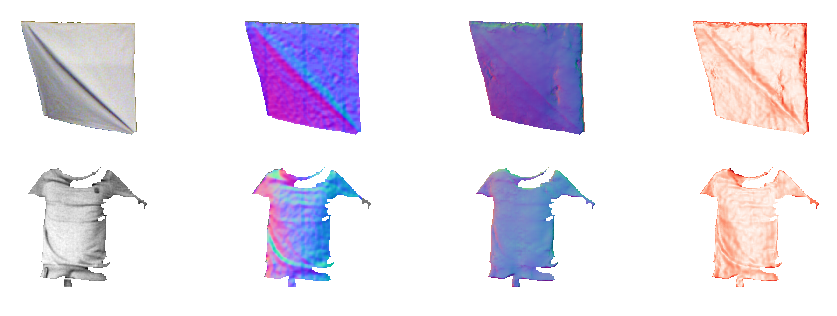}
{\textbf{Qualitative Results of Bednarik-Real Benchmark.} Visualization of the output on real objects from~\cite{bednarik2018learning} when trained on our synthetic data.\label{fig:qualitative_errors_real}}

\subsection{3D Reconstruction Benchmarks}

This section presents our 3D reconstruction benchmarks on the texture-less ShapeNet renders in our Synthetic (B) dataset using the 3D voxel grids from the ShapeNet dataset to benchmark monocular and multi-view reconstruction.

\noindent \textbf{Baselines.} \hspace{2pt} We choose networks known for their performance on the classic ShapeNet dataset and ones with available reproducible source code. For single-view reconstruction, we use Pix2Vox~\cite{xie2019pix2vox}, Pix2Vox++~\cite{xie2020pix2vox++}, and 3D-RETR~\cite{shi20213d}. Pix2Vox networks use convolutional frameworks, and 3D-RETR employs a Vision Transformer.

\noindent \textbf{Data Splits and Evaluation Metrics.} \hspace{2pt} We use the standard ShapeNet splits for training, validation, and testing. We train baseline networks on the training set and measure performance using the Intersection over Union (IoU) metric on the validation and test sets.

\subsubsection{Monocular Benchmark}
\label{sec:benchmark_monocular}

The Monocular benchmark assesses 3D reconstruction using a single RGB image. We train the models using only the first image for each training set model and reconstruct the $32^2$ voxels for the models in validation and test sets. Table~\ref{tab:results_svbenchmark} presents the results of this benchmark, with all three networks showing competitive performance. This indicates the ability of our texture-less renders to support 3D reconstruction even in a challenging monocular scenario.

\begin{table}[ht]
\centering
\scriptsize
\setlength{\tabcolsep}{3pt}
\caption{\textbf{Monocular Reconstruction Benchmark for Synthetic (B).} Results of the single-view reconstruction benchmark for the validation and test sets. The transformer-based 3D-RETR network shows the best performance.}
\label{tab:results_svbenchmark}
\begin{tabularx}{\linewidth}{|X|ccc|ccc|}
\hline
 & \multicolumn{3}{c|}{Validation} & \multicolumn{3}{c|}{Test} \\
 \hline
Object & 3D-RETR & Pix2Vox & Pix2Vox++ & 3D-RETR & Pix2Vox & Pix2Vox++ \\
\hline
plane   & \textbf{0.585} & 0.478 & 0.522 & \textbf{0.576} & 0.461 & 0.507 \\ 
bench   & \textbf{0.513} & 0.314 & 0.395 & \textbf{0.485} & 0.298 & 0.380 \\ 
cabinet & \textbf{0.654} & 0.512 & 0.605 & \textbf{0.650} & 0.515 & 0.607 \\ 
car     & \textbf{0.783} & 0.727 & 0.782 & \textbf{0.788} & 0.726 & 0.784 \\ 
chair   & \textbf{0.447} & 0.287 & 0.345 & \textbf{0.447} & 0.290 & 0.347 \\ 
display & \textbf{0.440} & 0.303 & 0.356 & \textbf{0.454} & 0.333 & 0.376 \\ 
lamp    & \textbf{0.410} & 0.291 & 0.298 & \textbf{0.400} & 0.291 & 0.301 \\
rifle   & \textbf{0.573} & 0.482 & 0.551 & \textbf{0.559} & 0.458 & 0.555 \\ 
sofa    & \textbf{0.640} & 0.480 & 0.519 & \textbf{0.624} & 0.474 & 0.514 \\ 
speaker & \textbf{0.608} & 0.518 & 0.557 & \textbf{0.641} & 0.497 & 0.542 \\ 
table   & \textbf{0.500} & 0.273 & 0.397 & \textbf{0.500} & 0.273 & 0.393 \\ 
phone   & \textbf{0.644} & 0.488 & 0.562 & \textbf{0.676} & 0.529 & 0.612 \\ 
watercraft & \textbf{0.521} & 0.387 & 0.467 & \textbf{0.513} & 0.363 & 0.461 \\ 
\hline
mean    & \textbf{0.563} & 0.403 & 0.471 & \textbf{0.563} & 0.397 & 0.468 \\ 
\hline
\end{tabularx}
\end{table}

\subsubsection{Multi-View Benchmark}
\label{sec:benchmark_multi-view}

In this benchmark, we evaluate the task of 3D reconstruction by leveraging multiple images of the same object. It provides a comprehensive assessment of the ability of networks to take advantage of multiple perspectives. To assess the robustness of the networks at various complexity levels, we present results for different numbers of views: namely 2, 5, and 8. 

\begin{figure}[ht]
\centering

\subfloat[Validation Set]{
    \subfloat{
        \includegraphics[clip,trim={0 21cm 0 0},width=0.31\linewidth]{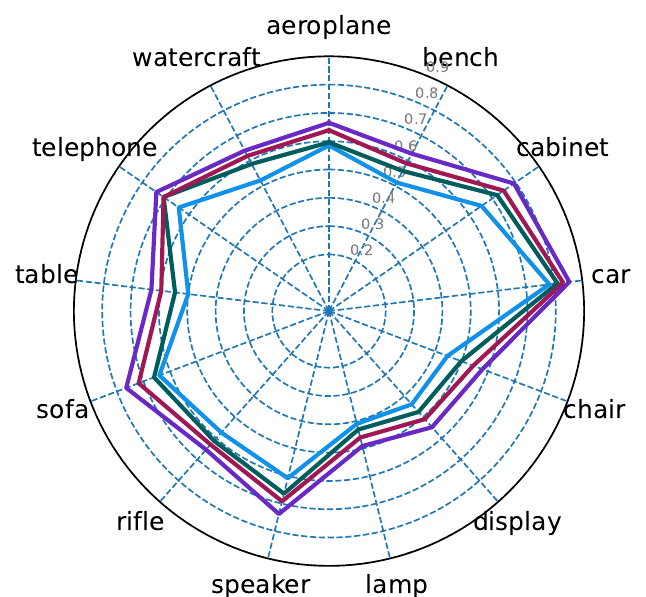}
    }
    \subfloat{
        \includegraphics[clip,trim={0 10.5cm 0 10.5cm},width=0.31\linewidth]{figures/results/iou_categorical_val_tless.pdf}
    }
    \subfloat{
        \includegraphics[clip,trim={0 0 0 21cm},width=0.31\linewidth]{figures/results/iou_categorical_val_tless.pdf}
    }
}

\subfloat[Test Set]{
    \subfloat{
        \includegraphics[clip,trim={0 21cm 0 0},width=0.31\linewidth]{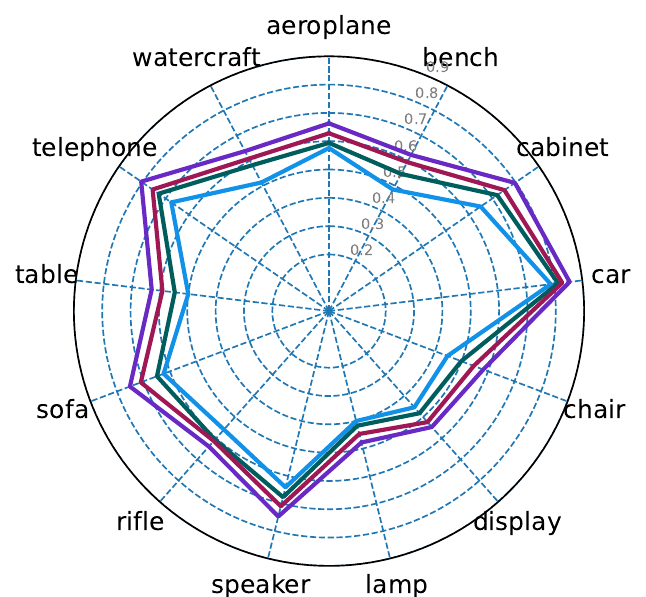}
    }
    \subfloat{
        \includegraphics[clip,trim={0 10.5cm 0 10.5cm},width=0.31\linewidth]{figures/results/iou_categorical_test_tless.pdf}
    }
    \subfloat{
        \includegraphics[clip,trim={0 0 0 21cm},width=0.31\linewidth]{figures/results/iou_categorical_test_tless.pdf}
    }
}

\caption{\textbf{Category-wise IoU for the val and test sets.} Comparison of 3D-RETR, Pix2Vox, and Pix2Vox++ using various numbers of input images on the validation and test sets. \textit{Best viewed in color and zoomed-in on a screen.}}
\label{fig:iou_comparison}
\end{figure}

Figure~\ref{fig:iou_comparison} illustrates the category-wise performance for each model, with complete results provided in Table~\ref{tab:results_mvbenchmark_textureless}.

\begin{table*}[ht]
\centering
\setlength{\tabcolsep}{4pt}
\caption{\textbf{Multi-View Reconstruction Benchmark.} Results of the multi-view reconstruction benchmark on validation and test sets with different training views.}
\label{tab:results_mvbenchmark_textureless}
\begin{tabularx}{\linewidth}{|cXcccccccccccccc|}
\hline
 Views & Method & plane & bench & cabinet & car & chair & display & lamp & rifle & sofa & speaker & table & phone & watercraft & mean \\
\hline
\multicolumn{16}{|c|}{Validation Set} \\
\hline
\multirow{3}{*}{1}
 & 3D-RETR   & \textbf{0.585} & \textbf{0.513} & \textbf{0.654} & \textbf{0.783} & \textbf{0.447} & \textbf{0.440} & \textbf{0.410} & \textbf{0.608} & \textbf{0.573} & \textbf{0.640} & \textbf{0.500} & \textbf{0.644} & \textbf{0.521} & \textbf{0.563} \\
 & Pix2Vox   & {0.446} & {0.270} & {0.468} & {0.704} & {0.222} & {0.253} & {0.250} & {0.421} & {0.434} & {0.494} & {0.221} & {0.508} & {0.346} & {0.359} \\
 & Pix2Vox++ & {0.502} & {0.344} & {0.579} & {0.776} & {0.282} & {0.317} & {0.252} & {0.520} & {0.489} & {0.541} & {0.351} & {0.545} & {0.435} & {0.435} \\
\hline
\multirow{3}{*}{2}
 & 3D-RETR   & \textbf{0.596} & \textbf{0.559} & \textbf{0.721} & \textbf{0.812} & \textbf{0.499} & \textbf{0.476} & \textbf{0.431} & \textbf{0.664} & \textbf{0.612} & \textbf{0.660} & \textbf{0.547} & \textbf{0.709} & \textbf{0.587} & \textbf{0.605} \\
 & Pix2Vox   & {0.482} & {0.328} & {0.604} & {0.710} & {0.301} & {0.305} & {0.237} & {0.522} & {0.505} & {0.525} & {0.316} & {0.545} & {0.456} & {0.422} \\
 & Pix2Vox++ & {0.528} & {0.435} & {0.648} & {0.788} & {0.349} & {0.335} & {0.280} & {0.570} & {0.546} & {0.582} & {0.413} & {0.583} & {0.493} & {0.484} \\
\hline
\multirow{3}{*}{5}
 & 3D-RETR   & \textbf{0.639} & \textbf{0.589} & \textbf{0.749} & \textbf{0.832} & \textbf{0.544} & \textbf{0.510} & \textbf{0.460} & \textbf{0.692} & \textbf{0.630} & \textbf{0.718} & \textbf{0.596} & \textbf{0.709} & \textbf{0.619} & \textbf{0.637} \\
 & Pix2Vox   & {0.506} & {0.397} & {0.604} & {0.728} & {0.344} & {0.342} & {0.286} & {0.529} & {0.537} & {0.550} & {0.395} & {0.623} & {0.489} & {0.466} \\
 & Pix2Vox++ & {0.612} & {0.508} & {0.701} & {0.812} & {0.452} & {0.415} & {0.362} & {0.626} & {0.588} & {0.655} & {0.511} & {0.654} & {0.577} & {0.563} \\
\hline
\multirow{2}{*}{8}
 & 3D-RETR   & \textbf{0.665} & \textbf{0.628} & \textbf{0.793} & \textbf{0.854} & \textbf{0.579} & \textbf{0.548} & \textbf{0.491} & \textbf{0.738} & \textbf{0.654} & \textbf{0.765} & \textbf{0.631} & \textbf{0.740} & \textbf{0.640} & \textbf{0.671} \\
 & Pix2Vox   & {0.556} & {0.423} & {0.647} & {0.786} & {0.399} & {0.363} & {0.318} & {0.553} & {0.549} & {0.598} & {0.427} & {0.621} & {0.534} & {0.505} \\
 % & Pix2Vox++ & {0.478} & {0.282} & {0.538} & {0.688} & {0.194} & {0.276} & {0.179} & {0.481} & {0.443} & {0.504} & {0.318} & {0.473} & {0.405} & {0.382} \\
\hline
\multicolumn{16}{|c|}{Test Set} \\
\hline
\multirow{3}{*}{1}
 & 3D-RETR   & \textbf{0.576} & \textbf{0.485} & \textbf{0.650} & \textbf{0.788} & \textbf{0.447} & \textbf{0.454} & \textbf{0.400} & \textbf{0.641} & \textbf{0.559} & \textbf{0.624} & \textbf{0.500} & \textbf{0.676} & \textbf{0.513} & \textbf{0.563} \\
 & Pix2Vox   & {0.431} & {0.245} & {0.469} & {0.703} & {0.226} & {0.301} & {0.254} & {0.394} & {0.438} & {0.473} & {0.221} & {0.548} & {0.332} & {0.355} \\
 & Pix2Vox++ & {0.487} & {0.335} & {0.574} & {0.776} & {0.284} & {0.339} & {0.259} & {0.513} & {0.483} & {0.522} & {0.345} & {0.605} & {0.432} & {0.431} \\
\hline
\multirow{3}{*}{2}
 & 3D-RETR   & \textbf{0.594} & \textbf{0.547} & \textbf{0.721} & \textbf{0.814} & \textbf{0.499} & \textbf{0.482} & \textbf{0.417} & \textbf{0.677} & \textbf{0.607} & \textbf{0.649} & \textbf{0.548} & \textbf{0.730} & \textbf{0.582} & \textbf{0.605} \\
 & Pix2Vox   & {0.474} & {0.303} & {0.597} & {0.715} & {0.305} & {0.323} & {0.236} & {0.494} & {0.488} & {0.506} & {0.320} & {0.604} & {0.441} & {0.419} \\
 & Pix2Vox++ & {0.519} & {0.418} & {0.640} & {0.787} & {0.352} & {0.355} & {0.284} & {0.561} & {0.540} & {0.560} & {0.406} & {0.660} & {0.493} & {0.481} \\
\hline
\multirow{3}{*}{5}
 & 3D-RETR   & \textbf{0.628} & \textbf{0.594} & \textbf{0.752} & \textbf{0.829} & \textbf{0.546} & \textbf{0.524} & \textbf{0.447} & \textbf{0.710} & \textbf{0.613} & \textbf{0.710} & \textbf{0.592} & \textbf{0.756} & \textbf{0.608} & \textbf{0.639} \\
 & Pix2Vox   & {0.485} & {0.393} & {0.631} & {0.730} & {0.359} & {0.375} & {0.285} & {0.522} & {0.532} & {0.552} & {0.396} & {0.675} & {0.477} & {0.467} \\
 & Pix2Vox++ & {0.606} & {0.500} & {0.720} & {0.815} & {0.457} & {0.435} & {0.362} & {0.617} & {0.592} & {0.650} & {0.507} & {0.705} & {0.571} & {0.562} \\
\hline
\multirow{3}{*}{8}
 & 3D-RETR   & \textbf{0.663} & \textbf{0.623} & \textbf{0.797} & \textbf{0.855} & \textbf{0.582} & \textbf{0.547} & \textbf{0.478} & \textbf{0.747} & \textbf{0.639} & \textbf{0.751} & \textbf{0.630} & \textbf{0.805} & \textbf{0.638} & \textbf{0.673} \\
 & Pix2Vox   & {0.538} & {0.412} & {0.654} & {0.791} & {0.404} & {0.396} & {0.312} & {0.559} & {0.543} & {0.586} & {0.431} & {0.693} & {0.527} & {0.505} \\
 & Pix2Vox++ & {0.596} & {0.519} & {0.755} & {0.836} & {0.479} & {0.442} & {0.358} & {0.694} & {0.592} & {0.696} & {0.526} & {0.728} & {0.570} & {0.582} \\
\hline
\end{tabularx}
\end{table*}

\section{Conclusion}
\label{sec:concluding_remarks}

This work introduces the first large-scale depth and surface normals dataset dedicated to texture-less surfaces, marking a significant step forward in 3D shape analysis. The dataset contains depth and surface normal maps, providing a 2.5D representation of object shapes. While this offers partial geometry from a single viewpoint, it sets a foundation for reconstructing texture-less surfaces—a nascent but growing area of research.

The choice of synthetic data over real-world data addresses the complexity of large-scale 3D data acquisition. We align our normal maps computation with existing methodologies for consistency while offering an avenue for generating more accurate normals using Blender's capabilities. This flexibility supports future work in synthetic data generation that may require higher fidelity in surface normal representation.

Our dataset surpasses existing texture-less datasets in size and diversity, offering a new benchmark for texture-less surface reconstruction methods. Future research can leverage this dataset to test the scalability and generalizability of existing approaches. Additionally, providing source code enables the community to expand the dataset further. Such work could significantly advance shape understanding and reconstruction techniques, particularly for texture-less objects.

\appendices

\section*{Ethical Considerations}
We acknowledge the inclusion of cultural and religious symbols in our 3D models. All images are rendered respectfully, ensuring the depictions remain neutral and free from offensive contexts.

\section*{Acknowledgment}
We extend our gratitude to the ShapeNet repository contributors for providing the 3D models that partly form the basis of our dataset and to the Blender developers for the software that enabled our renderings.

\bibliographystyle{unsrt}
\bibliography{main}

\begin{thebibliography}{10}

\bibitem{wang2018pixel2mesh}
Nanyang Wang, Yinda Zhang, Zhuwen Li, Yanwei Fu, Wei Liu, and Yu-Gang Jiang.
\newblock Pixel2mesh: Generating 3d mesh models from single rgb images.
\newblock In {\em Proceedings of the European conference on computer vision (ECCV)}, pages 52--67, 2018.

\bibitem{bednarik2018learning}
Jan Bednarik, Pascal Fua, and Mathieu Salzmann.
\newblock Learning to reconstruct texture-less deformable surfaces from a single view.
\newblock In {\em 2018 International Conference on 3D Vision (3DV)}, pages 606--615. IEEE, 2018.

\bibitem{golyanik2018hdm}
Vladislav Golyanik, Soshi Shimada, Kiran Varanasi, and Didier Stricker.
\newblock Hdm-net: Monocular non-rigid 3d reconstruction with learned deformation model.
\newblock {\em CoRR}, abs/1803.10193, 2018.

\bibitem{shimada2019ismo}
Soshi Shimada, Vladislav Golyanik, Christian Theobalt, and Didier Stricker.
\newblock Ismo-gan: Adversarial learning for monocular non-rigid 3d reconstruction.
\newblock {\em CoRR}, abs/1904.12144, 2019.

\bibitem{tsoli2019patch}
Aggeliki Tsoli and Antonis.~A. Argyros.
\newblock Patch-based reconstruction of a textureless deformable 3d surface from a single rgb image.
\newblock In {\em Proceedings of the IEEE/CVF International Conference on Computer Vision (ICCV) Workshops}, October 2019.

\bibitem{li2021vrvt}
Xi~Li and Ping Kuang.
\newblock 3d-vrvt: 3d voxel reconstruction from a single image with vision transformer.
\newblock In {\em 2021 International Conference on Culture-oriented Science \& Technology (ICCST)}, pages 343--348. IEEE, 2021.

\bibitem{ley2016reconstructing}
Andreas Ley, Ronny H{\"a}nsch, and Olaf Hellwich.
\newblock Reconstructing white walls: Multi-view, multi-shot 3d reconstruction of textureless surfaces.
\newblock {\em ISPRS Annals of Photogrammetry, Remote Sensing \& Spatial Information Sciences}, 3(3), 2016.

\bibitem{wang2016template}
Xuan Wang, Mathieu Salzmann, Fei Wang, and Jizhong Zhao.
\newblock Template-free 3d reconstruction of poorly-textured nonrigid surfaces.
\newblock In {\em Computer Vision--ECCV 2016: 14th European Conference, Amsterdam, The Netherlands, October 11--14, 2016, Proceedings, Part VII 14}, pages 648--663. Springer, 2016.

\bibitem{hafeez2017image}
Jahanzeb Hafeez, Seunghyun Lee, Soonchul Kwon, and Alaric Hamacher.
\newblock Image based 3d reconstruction of texture-less objects for vr contents.
\newblock {\em International journal of advanced smart convergence}, 6(1):9--17, 2017.

\bibitem{ahmadabadian2019automatic}
Ali~Hosseininaveh Ahmadabadian, Ali Karami, and Rouhallah Yazdan.
\newblock An automatic 3d reconstruction system for texture-less objects.
\newblock {\em Robotics and Autonomous Systems}, 117:29--39, 2019.

\bibitem{santovsi2019evaluation}
{\v{Z}}eljko Santo{\v{s}}i, Igor Budak, Vesna Stojakovi{\'c}, Mario {\v{S}}okac, and Dorde Vukeli{\'c}.
\newblock Evaluation of synthetically generated patterns for image-based 3d reconstruction of texture-less objects.
\newblock {\em Measurement}, 147:106883, 2019.

\bibitem{fan20213d}
Jiacheng Fan, Yuan Feng, Jinqiu Mo, Shigang Wang, and Qinghua Liang.
\newblock 3d reconstruction of non-textured surface by combining shape from shading and stereovision.
\newblock {\em Measurement}, 185:110029, 2021.

\bibitem{cheng2021multi}
Ziang Cheng, Hongdong Li, Yuta Asano, Yinqiang Zheng, and Imari Sato.
\newblock Multi-view 3d reconstruction of a texture-less smooth surface of unknown generic reflectance.
\newblock In {\em Proceedings of the IEEE/CVF Conference on Computer Vision and Pattern Recognition}, pages 16226--16235, 2021.

\bibitem{yang2021practical}
Xuyuan Yang and Guang Jiang.
\newblock A practical 3d reconstruction method for weak texture scenes.
\newblock {\em Remote Sensing}, 13(16):3103, 2021.

\bibitem{ye2023self}
Botao Ye, Sifei Liu, Xueting Li, and Ming-Hsuan Yang.
\newblock Self-supervised super-plane for neural 3d reconstruction.
\newblock In {\em Proceedings of the IEEE/CVF Conference on Computer Vision and Pattern Recognition}, pages 21415--21424, 2023.

\bibitem{hodan2017t}
Tom{\'a}{\v{s}} Hodan, Pavel Haluza, {\v{S}}tep{\'a}n Obdr{\v{z}}{\'a}lek, Jiri Matas, Manolis Lourakis, and Xenophon Zabulis.
\newblock T-less: An rgb-d dataset for 6d pose estimation of texture-less objects.
\newblock In {\em 2017 IEEE Winter Conference on Applications of Computer Vision (WACV)}, pages 880--888. IEEE, 2017.

\bibitem{widya2019whole}
Aji~Resindra Widya, Yusuke Monno, Masatoshi Okutomi, Sho Suzuki, Takuji Gotoda, and Kenji Miki.
\newblock Whole stomach 3d reconstruction and frame localization from monocular endoscope video.
\newblock {\em IEEE Journal of Translational Engineering in Health and Medicine}, 7:1--10, 2019.

\bibitem{boven2020diagnostic}
Judith B{\"o}ven, Johannes Boos, Andrea Steuwe, Janna Morawitz, Lino~Morris Sawicki, Julian Caspers, Lisa K{\"u}ppers, Benno Hartung, Christoph Thomas, Gerald Antoch, et~al.
\newblock Diagnostic value and forensic relevance of a novel photorealistic 3d reconstruction technique in post-mortem ct.
\newblock {\em The British Journal of Radiology}, 93(1112):20200204, 2020.

\bibitem{sun2017revisiting}
Chen Sun, Abhinav Shrivastava, Saurabh Singh, and Abhinav Gupta.
\newblock Revisiting unreasonable effectiveness of data in deep learning era.
\newblock In {\em Proceedings of the IEEE international conference on computer vision}, pages 843--852, 2017.

\bibitem{chang2015shapenet}
Angel~X. Chang, Thomas Funkhouser, Leonidas Guibas, Pat Hanrahan, Qixing Huang, Zimo Li, Silvio Savarese, Manolis Savva, Shuran Song, Hao Su, Jianxiong Xiao, Li~Yi, and Fisher Yu.
\newblock Shapenet: An information-rich 3d model repository.
\newblock Technical Report arXiv:1512.03012 [cs.GR], Stanford University --- Princeton University --- Toyota Technological Institute at Chicago, 2015.

\bibitem{sun2018pix3d}
Xingyuan Sun, Jiajun Wu, Xiuming Zhang, Zhoutong Zhang, Chengkai Zhang, Tianfan Xue, Joshua~B Tenenbaum, and William~T Freeman.
\newblock Pix3d: Dataset and methods for single-image 3d shape modeling.
\newblock In {\em Proceedings of the IEEE conference on computer vision and pattern recognition}, pages 2974--2983, 2018.

\bibitem{xie2020pix2vox++}
Haozhe Xie, Hongxun Yao, Shengping Zhang, Shangchen Zhou, and Wenxiu Sun.
\newblock Pix2vox++: Multi-scale context-aware 3d object reconstruction from single and multiple images.
\newblock {\em International Journal of Computer Vision}, 128(12):2919--2935, 2020.

\bibitem{khan2022three}
Muhammad Saif~Ullah Khan, Alain Pagani, Marcus Liwicki, Didier Stricker, and Muhammad~Zeshan Afzal.
\newblock Three-dimensional reconstruction from a single rgb image using deep learning: A review.
\newblock {\em Journal of Imaging}, 8(9), 2022.

\bibitem{godard2019digging}
Cl{\'e}ment Godard, Oisin Mac~Aodha, Michael Firman, and Gabriel~J Brostow.
\newblock Digging into self-supervised monocular depth estimation.
\newblock In {\em Proceedings of the IEEE/CVF international conference on computer vision}, pages 3828--3838, 2019.

\bibitem{costanzino2023learning}
Alex Costanzino, Pierluigi~Zama Ramirez, Matteo Poggi, Fabio Tosi, Stefano Mattoccia, and Luigi Di~Stefano.
\newblock Learning depth estimation for transparent and mirror surfaces.
\newblock In {\em Proceedings of the IEEE/CVF International Conference on Computer Vision}, pages 9244--9255, 2023.

\bibitem{spencer2023monocular}
Jaime Spencer, C~Stella Qian, Chris Russell, Simon Hadfield, Erich Graf, Wendy Adams, Andrew~J Schofield, James~H Elder, Richard Bowden, Heng Cong, et~al.
\newblock The monocular depth estimation challenge.
\newblock In {\em Proceedings of the IEEE/CVF Winter Conference on Applications of Computer Vision}, pages 623--632, 2023.

\bibitem{gasperini2023robust}
Stefano Gasperini, Nils Morbitzer, HyunJun Jung, Nassir Navab, and Federico Tombari.
\newblock Robust monocular depth estimation under challenging conditions.
\newblock In {\em Proceedings of the IEEE/CVF international conference on computer vision}, pages 8177--8186, 2023.

\bibitem{ming2021deep}
Yue Ming, Xuyang Meng, Chunxiao Fan, and Hui Yu.
\newblock Deep learning for monocular depth estimation: A review.
\newblock {\em Neurocomputing}, 438:14--33, 2021.

\bibitem{lenssen2020deep}
Jan~Eric Lenssen, Christian Osendorfer, and Jonathan Masci.
\newblock Deep iterative surface normal estimation.
\newblock In {\em Proceedings of the ieee/cvf conference on computer vision and pattern recognition}, pages 11247--11256, 2020.

\bibitem{klasing2009comparison}
Klaas Klasing, Daniel Althoff, Dirk Wollherr, and Martin Buss.
\newblock Comparison of surface normal estimation methods for range sensing applications.
\newblock In {\em 2009 IEEE international conference on robotics and automation}, pages 3206--3211. Ieee, 2009.

\bibitem{wang2015designing}
Xiaolong Wang, David Fouhey, and Abhinav Gupta.
\newblock Designing deep networks for surface normal estimation.
\newblock In {\em Proceedings of the IEEE conference on computer vision and pattern recognition}, pages 539--547, 2015.

\bibitem{wu2017marrnet}
Jiajun Wu, Yifan Wang, Tianfan Xue, Xingyuan Sun, Bill Freeman, and Josh Tenenbaum.
\newblock Marrnet: 3d shape reconstruction via 2.5 d sketches.
\newblock {\em Advances in neural information processing systems}, 30, 2017.

\bibitem{yuniarti2019review}
Anny Yuniarti and Nanik Suciati.
\newblock A review of deep learning techniques for 3d reconstruction of 2d images.
\newblock In {\em 2019 12th International Conference on Information \& Communication Technology and System (ICTS)}, pages 327--331. IEEE, 2019.

\bibitem{maxim2021survey}
Bogdan Maxim and Sergiu Nedevschi.
\newblock A survey on the current state of the art on deep learning 3d reconstruction.
\newblock In {\em 2021 IEEE 17th International Conference on Intelligent Computer Communication and Processing (ICCP)}, pages 283--290. IEEE, 2021.

\bibitem{mittal2022autosdf}
Paritosh Mittal, Yen-Chi Cheng, Maneesh Singh, and Shubham Tulsiani.
\newblock Autosdf: Shape priors for 3d completion, reconstruction and generation.
\newblock In {\em Proceedings of the IEEE/CVF Conference on Computer Vision and Pattern Recognition}, pages 306--315, 2022.

\bibitem{wen20223d}
Xin Wen, Junsheng Zhou, Yu-Shen Liu, Hua Su, Zhen Dong, and Zhizhong Han.
\newblock 3d shape reconstruction from 2d images with disentangled attribute flow.
\newblock In {\em Proceedings of the IEEE/CVF conference on computer vision and pattern recognition}, pages 3803--3813, 2022.

\bibitem{alwala2022pre}
Kalyan~Vasudev Alwala, Abhinav Gupta, and Shubham Tulsiani.
\newblock Pre-train, self-train, distill: A simple recipe for supersizing 3d reconstruction.
\newblock In {\em Proceedings of the IEEE/CVF Conference on Computer Vision and Pattern Recognition}, pages 3773--3782, 2022.

\bibitem{tang2021skeletonnet}
Jiapeng Tang, Xiaoguang Han, Mingkui Tan, Xin Tong, and Kui Jia.
\newblock Skeletonnet: A topology-preserving solution for learning mesh reconstruction of object surfaces from rgb images.
\newblock {\em IEEE transactions on pattern analysis and machine intelligence}, 44(10):6454--6471, 2021.

\bibitem{zhang2020training}
Biao Zhang and Peter Wonka.
\newblock Training data generating networks: Shape reconstruction via bi-level optimization.
\newblock {\em arXiv preprint arXiv:2010.08276}, 2020.

\bibitem{xing2022few}
Zhen Xing, Yijiang Chen, Zhixin Ling, Xiangdong Zhou, and Yu~Xiang.
\newblock Few-shot single-view 3d reconstruction with memory prior contrastive network.
\newblock In {\em European Conference on Computer Vision}, pages 55--70. Springer, 2022.

\bibitem{xing2022semi}
Zhen Xing, Hengduo Li, Zuxuan Wu, and Yu-Gang Jiang.
\newblock Semi-supervised single-view 3d reconstruction via prototype shape priors.
\newblock In {\em European Conference on Computer Vision}, pages 535--551. Springer, 2022.

\bibitem{chen2023single}
Yixin Chen, Junfeng Ni, Nan Jiang, Yaowei Zhang, Yixin Zhu, and Siyuan Huang.
\newblock Single-view 3d scene reconstruction with high-fidelity shape and texture.
\newblock {\em arXiv preprint arXiv:2311.00457}, 2023.

\bibitem{choy20163d}
Christopher~B Choy, Danfei Xu, JunYoung Gwak, Kevin Chen, and Silvio Savarese.
\newblock 3d-r2n2: A unified approach for single and multi-view 3d object reconstruction.
\newblock In {\em Proceedings of the European Conference on Computer Vision ({ECCV})}, 2016.

\bibitem{tiong20223d}
Leslie Ching~Ow Tiong, Dick Sigmund, and Andrew Beng~Jin Teoh.
\newblock 3d-c2ft: Coarse-to-fine transformer for multi-view 3d reconstruction.
\newblock In {\em Proceedings of the Asian Conference on Computer Vision}, pages 1438--1454, 2022.

\bibitem{yang2022fvor}
Zhenpei Yang, Zhile Ren, Miguel~Angel Bautista, Zaiwei Zhang, Qi~Shan, and Qixing Huang.
\newblock Fvor: Robust joint shape and pose optimization for few-view object reconstruction.
\newblock In {\em Proceedings of the IEEE/CVF Conference on Computer Vision and Pattern Recognition}, pages 2497--2507, 2022.

\bibitem{xie2019pix2vox}
Haozhe Xie, Hongxun Yao, Xiaoshuai Sun, Shangchen Zhou, and Shengping Zhang.
\newblock Pix2vox: Context-aware 3d reconstruction from single and multi-view images.
\newblock In {\em Proceedings of the IEEE/CVF international conference on computer vision}, pages 2690--2698, 2019.

\bibitem{shi20213d}
Zai Shi, Zhao Meng, Yiran Xing, Yunpu Ma, and Roger Wattenhofer.
\newblock 3d-retr: end-to-end single and multi-view 3d reconstruction with transformers.
\newblock {\em arXiv preprint arXiv:2110.08861}, 2021.

\bibitem{yuan2021vanet}
Yi~Yuan, Jilin Tang, and Zhengxia Zou.
\newblock Vanet: a view attention guided network for 3d reconstruction from single and multi-view images.
\newblock In {\em 2021 IEEE International Conference on Multimedia and Expo (ICME)}, pages 1--6, 2021.

\bibitem{community2018blender}
Blender~Online Community.
\newblock {\em Blender - a 3D modelling and rendering package}.
\newblock Blender Foundation, Stichting Blender Foundation, Amsterdam, 2018.

\bibitem{wasenmuller2016comparison}
Oliver Wasenm{\"u}ller and Didier Stricker.
\newblock Comparison of kinect v1 and v2 depth images in terms of accuracy and precision.
\newblock In {\em Asian Conference on Computer Vision}, pages 34--45. Springer, 2016.

\bibitem{bertalmio2001navier}
Marcelo Bertalmio, Andrea~L Bertozzi, and Guillermo Sapiro.
\newblock Navier-stokes, fluid dynamics, and image and video inpainting.
\newblock In {\em Proceedings of the 2001 IEEE Computer Society Conference on Computer Vision and Pattern Recognition. CVPR 2001}, volume~1, pages I--I. IEEE, 2001.

\end{thebibliography}

\begin{IEEEbiography}[{\includegraphics[width=1in,height=1.25in,clip,keepaspectratio]{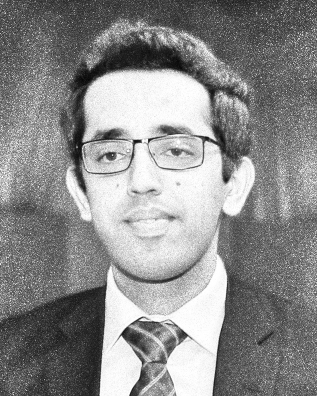}}]{Muhammad Saif Ullah Khan} earned a Bachelor's in software engineering from National University of Sciences and Technology (NUST), Islamabad, Pakistan, in 2018 and a Master's in computer science from Technical University of Kaiserslautern (TUK), Kaiserslautern, Germany, in 2022. He is a PhD candidate at the University of Kaiserslautern-Landau (RPTU), Kaiserslautern, Germany since January 2023.

He works as a Researcher in the Augmented Vision Group at the German Research Center for Artificial Intelligence (DFKI) in Kaiserslautern, Germany. His current research focuses on tracking human body motion from visual input and training multimodal foundation models for situational awareness. He has also worked on shape reconstruction, financial document understanding, and signature verification.
\end{IEEEbiography}

\begin{IEEEbiography}[{\includegraphics[width=1in,height=1.25in,clip,keepaspectratio]{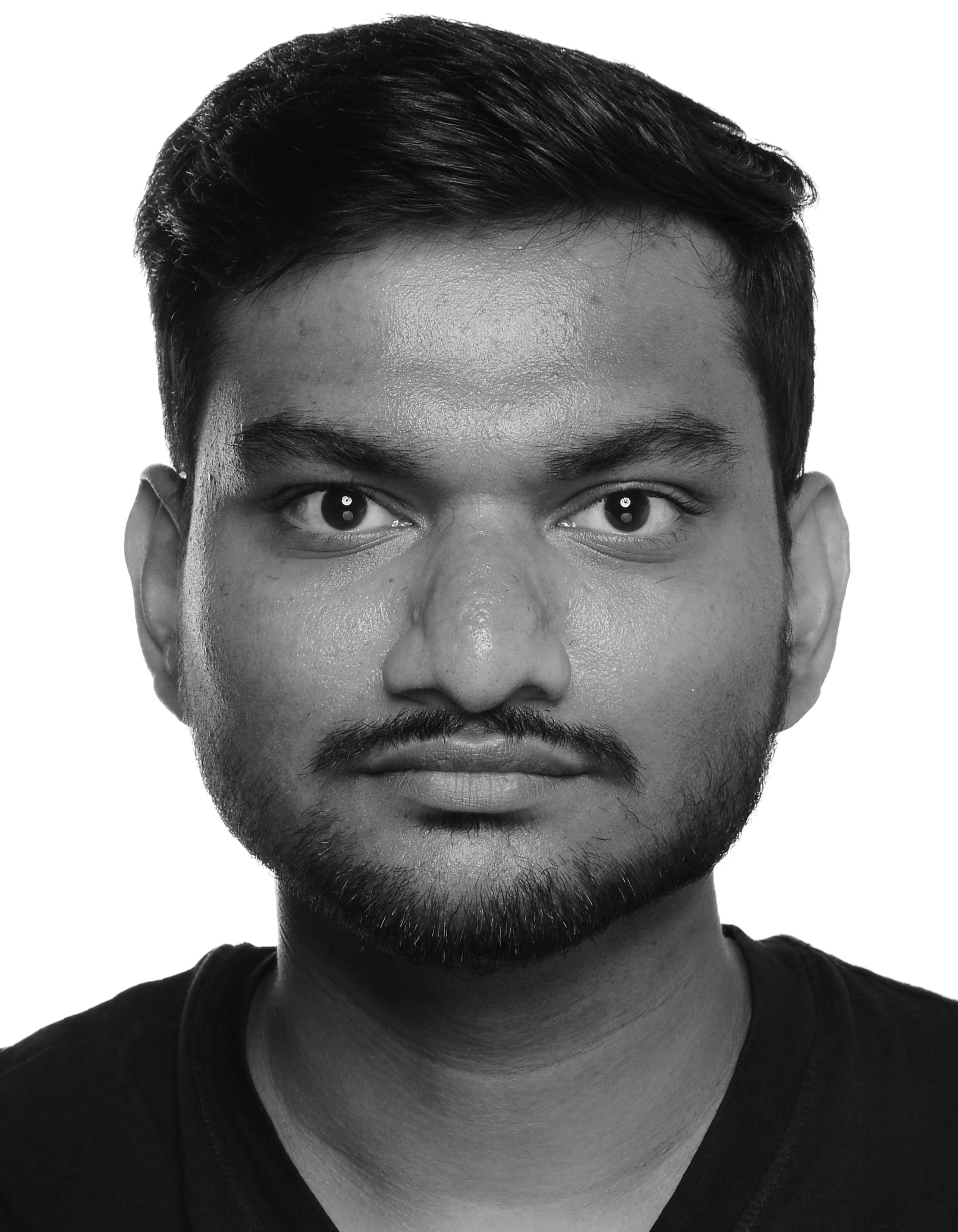}}]{Sankalp Sinha} is a Computer Science PhD candidate at RPTU in Germany, where he previously completed his Master's degree. He earned his Bachelor's in Computer Science and Engineering from SRM Institute of Science and Technology in India. Currently, Sankalp is a researcher at the German Research Center for Artificial Intelligence (DFKI), tackling AI-driven challenges.
 
His research interests are wide-ranging across AI. Sankalp is passionate about pushing the boundaries of document image analysis, particularly through zero-shot learning techniques that can handle unfamiliar data. He's also deeply involved in bridging visual and textual information, working with both vision-language models and large language models. Another area of focus is 3D reconstruction, where he explores innovative approaches, including text-driven parametrized 3D modeling. Through his work, he aims to create AI solutions that are not only cutting-edge but also practical and impactful.
\end{IEEEbiography}

\begin{IEEEbiography}[{\includegraphics[width=1in,height=1.25in,clip,keepaspectratio]{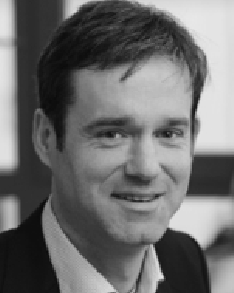}}]{Didier Stricker} led the Department of Virtual and Augmented Reality, Fraunhofer Institute for Computer Graphics, Darmstadt, Germany, from 2002 to 2008. He is currently a Professor at the Department of Computer Science, Rheinland-Pfälzische Technische Universität Kaiserslautern-Landau (RPTU), Germany. He is also a Scientific Director of the German Research Center for Artificial Intelligence (DFKI), Kaiserslautern, where he leads the Augmented Vision Research Group. His research interests include 3D computer vision, autonomous driving, wearable health, augmented reality applications, and deep learning. He received the Innovation Prize from the German Society of Computer Science in 2006. He serves as a reviewer for noteworthy journals in the area of VR/AR and computer vision.
\end{IEEEbiography}

\begin{IEEEbiography}[{\includegraphics[width=1in,height=1.25in,clip,keepaspectratio]{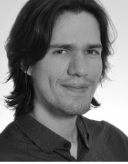}}]{Marcus Liwicki} (Senior Member, IEEE) received his M.S. in computer science from the Free University of Berlin, Germany, in 2004, his Ph.D. from the University of Bern, Switzerland, in 2007, and his Habilitation from the Technical University of Kaiserslautern, Germany, in 2011. He is currently a Chaired Professor at Luleå University of Technology and serves as a Senior Assistant at the University of Fribourg.

His research spans machine learning, pattern recognition, artificial intelligence, human–computer interaction, and document analysis. He has served as an Editor or Regular Reviewer for several prominent journals, including IEEE Transactions on Pattern Analysis and Machine Intelligence, IEEE Transactions on Audio, Speech, and Language Processing, International Journal of Document Analysis and Recognition, Frontiers of Computer Science, Frontiers in Digital Humanities, Pattern Recognition, and Pattern Recognition Letters.

Dr. Liwicki is a member of the International Association for Pattern Recognition (IAPR) and currently serves as Vice President of the Technical Committee 6. He also sits on the Governing Board of the International Graphonomics Society. He has chaired international workshops on Automated Forensic Handwriting Analysis and Document Analysis Systems (2014) and serves regularly on program committees for conferences in computer vision, pattern recognition, and machine learning.
\end{IEEEbiography}

\begin{IEEEbiography}[{\includegraphics[width=1in,height=1.25in,clip,keepaspectratio]{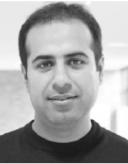}}]{Muhammad Zeshan Afzal} received the master’s degree majoring in visual computing from the University of Saarland, Germany, in 2010, and the Ph.D. degree majoring in artificial intelligence from the University of Technology, Kaiserslautern, Germany, in 2016.

His research interests include deep learning for vision and language understanding using deep learning. At an application level, his experiences include a generic segmentation framework for natural and human activity recognition, document and medical image analysis, scene text detection and recognition, as well as online and offline gesture recognition. Moreover, he is interested in recurrent neural networks for sequence processing applied to images and videos. He also worked with numerics for tensor-valued images. He worked in the industry (deep learning and AI lead insiders technologies GmbH) and academia (TU Kaiserslautern). He received the gold medal for the best graduating student in computer science from IUB
Pakistan, in 2002, and secured a DAAD (Germany) fellowship, in 2007. Dr. Afzal is a member of IAPR.
\end{IEEEbiography}

\EOD

\end{document}